\pdfoutput=1
\documentclass[table,10pt]{article} %
\usepackage{natbib,multicol}

\usepackage[frozencache,cachedir=.]{minted}
\usepackage{graphicx,url} %
\usepackage[algo2e,ruled,vlined,linesnumbered]{algorithm2e}
\usepackage{amsmath,amsthm,amssymb,url}
\usepackage{booktabs}
\usepackage{placeins,cleveref}
\usepackage{multicol}
\usepackage{tabularx,longtable}
\usepackage{comment}
\usepackage{multirow}
\usepackage{ifthen}
\DeclareMathOperator{\Bin}{Bin}
\DeclareMathOperator{\mut}{\textsc{Mutate}}
\DeclareSymbolFont{largesymbolsA}{U}{txexa}{m}{n}
\DeclareMathSymbol{\varprod}{\mathop}{largesymbolsA}{16}

\newcommand{\assign}{\leftarrow}
\newcommand{\fr}[1]{{\color{blue}NO MORE COMMENTS INSIDE THE TEX}}
\newcommand{\tb}[1]{{\color{violet}NO MORE COMMENTS INSIDE THE TEX}}
\newcommand{\mz}[1]{{\color{magenta}NO MORE COMMENTS INSIDE THE TEX}}
\newcommand{\lna}[1]{{\color{red}NO MORE COMMENTS INSIDE THE TEX}}

\def\lognormal{log-normal}
\def\Lognormal{Log-normal}

\title{\Lognormal{} Mutations and their Use in Detecting Surreptitious Fake Images}

\author{
Ismail Labiad\\   \url{ilabiad@meta.com},
       Meta AI Research 
      \and
      Thomas B\"ack \\   \url{t.h.w.baeck@liacs.leidenuniv.nl},
     Leiden University, LIACS,\\ Einsteinweg 55, 2333 CC Leiden, Netherlands
    \and
      Pierre Fernandez\\
      \url{pfz@meta.com},
       Meta AI Research 
\and
Laurent Najman\\
 \url{laurent.najman@esiee.fr},
 Univ Gustave Eiffel, CNRS, LIGM,\\ F-77454, Marne-la-Vall\'ee, France
      \and
Tom Sander \\
\url{tomsander@meta.com},
       Meta AI Research       
      \and
Furong Ye\\   \url{f.ye@liacs.leidenuniv.nl},
     Leiden University, \\ LIACS, Einsteinweg 55, 2333 CC Leiden, Netherlands
    \and
Mariia Zameshina\\
 \url{mariia.zameshina@esiee.fr},
  Univ Gustave Eiffel, CNRS, LIGM, \\F-77454, Marne-la-Vall\'ee, France
\and
       Olivier Teytaud \\ \url{oteytaud@meta.com},
       Meta AI Research }

\begin{document}
\date{}
\maketitle

\begin{abstract}
In many cases, adversarial attacks are based on specialized algorithms specifically dedicated to attacking automatic image classifiers. 
These algorithms perform well, thanks to an excellent ad hoc distribution of initial attacks. 
However, these attacks are easily detected due to their specific initial distribution.
We therefore consider other black-box attacks, inspired from generic black-box optimization tools, and in particular the \lognormal{} algorithm. 

We apply the \lognormal{} method to the attack of fake detectors, and get successful attacks: importantly, these attacks are not detected by detectors specialized on classical adversarial attacks.
Then, combining these attacks and deep detection, we create improved fake detectors.
\end{abstract}

\newcommand{\G}{Fake1}
\newcommand{\BG}{Fake2}
\newcommand{\PG}{Fake3}
\newcommand{\LDM}{LatentFake}
\newcommand{\IN}{RealWorld}

\section{Introduction and outline}

Adversarial attacks (such as Square Attack, SA~\citep{ACFH2020square}) or no-box attacks such as purifiers~\citep{noboxattack} can successfully imperceptibly modify fake images so that they are not recognized as fake by some fake detectors.  
Therefore, detecting these attacks is necessary for enhancing our fake detectors.
In the present paper, we analyze lesser known attacks inspired by the black-box optimization community: their algorithms perform well, and are sufficiently different for being undetected by detectors trained on classical attacks only.
\footnote{
All data collection and experiments were conducted on Jean Zay servers}
Iterative optimization heuristics such as Evolutionary Algorithms (EAs) have shown their strengths in solving hard optimization problems over the past decades. While various methods have been developed for optimization problems in particular domains, for instance, Genetic Algorithms and Evolution Strategies (ES) have been favored for discrete and continuous optimization, respectively, the variants inheriting their techniques have been proposed and applied across different domains. Regarding the topic of \emph{borrowing ideas across different domains}, there exists much work in exchanging ideas between discrete and continuous optimization domains. For example, variants of Covariance Matrix Adaptation ES (CMA-ES) and Differential Evolution (DE) have been proposed for specific problems~\citep{cma,HamanoSNS22,DasMS16}.
More recently, a study has investigated the ways of discretizing CMA-ES and its performance on different discrete BBOB problems~\citep{ThomaserNVYBK23}.
In this work, we {investigate the advantage of borrowing an idea from a self-adaptive pseudo-Boolean optimization algorithm to the continuous domain.}
One of EAs' key algorithmic components is the utilization of a probability distribution for generating new search points, including an iterative adaptation of this distribution based on the objective function values of the newly generated solutions \citep{DOERR2020106027,2020DN}.
A wide range of such update strategies has been proposed over the past decades, for example, for multivariate Gaussian distributions in the context of covariance matrix adaptation in evolution strategies for continuous optimization problems \citep{cma} and for binomial and a variety of other mutation strength distributions in the context of mutation rates in evolutionary algorithms for pseudo-Boolean optimization \citep{DOERR2020106027}.
Among those adaptation methods for the mutation strength in pseudo-Boolean optimization, the so-called \textbf{log-normal mutation} was developed almost thirty years ago as a method for allowing the mutation strength to quickly shift from exploration to exploitation \citep{Baeck1995,Baeck1996lognormal,Kruisselbrink2011lognormal}. 
As shown in \citep{DOERR2020106027}, on a set of 23 pseudo-Boolean test functions, \lognormal{} mutation shows an empirical cumulative distribution functions (ECDF) performance across all 23 functions that are very close to the best-performing algorithm.
In this paper, we  compare the \lognormal{} (LN) mutation  
 
with various algorithms for continuous optimization problems selected from the Nevergrad benchmarks~\citep{nevergrad}.  Experimental results indicate that the \lognormal{}  strategy is competitive for problems that are %
particularly difficult, i.e., multi-modal and highly-deceptive. %
Inspired by the mentioned benchmarking results, we test the \lognormal{} mutation for a practical scenario, namely 

the attack of fake detectors.

{\bf{Outline:}} Section \ref{sec:sota} presents the state of the art in Black-Box Optimization 
 (BBO, \cref{sec:bbo}) and fake detectors (\cref{sec:fd}).
Section \ref{sec:met} present our tools: \cref{sec:alg} for the \lognormal{} mutations, \cref{sec:alg:extend}
 for the extension of \lognormal{} to  continuous domains.
 
 Then \cref{sec:experiments} presents experimental results, including an ablation in \cref{lognormcompa}, results on the Nevergrad benchmark suite~(\Cref{sec:nev}), discussing specific results (\Cref{selected}), 
 and the application to fake detectors (\Cref{sec:fake}).
 Comparing various algorithms on this application, we observe particularly good performance of \lognormal{} mutation.
 
{\bf{Contributions:}}
(1) We present an extension of  \lognormal{} mutation (usually applied in the context of discrete domains)  
    to arbitrary discrete, continuous and mixed search spaces (Section \ref{sec:alg:extend}).
    (2) We present a comparison between \lognormal{}mutation and a state-of-the-art algorithm selector~\citep{NGOptMeunierTEVC22}, illustrating a particular strength of \lognormal{} for deceptive and multimodal problems (Section \ref{sec:nev}, \ref{selected}). %
    (3) We show that \lognormal{} mutation is particularly well suited for attacking fake detectors: \lognormal{} mutation performs well (Section \ref{perf}), and the created fake images go undetected by detectors based on existing attacks (Section \ref{sa_detection}). For converting the attacks into defense, we then add a detector of these attacks (Section  \ref{ln_detector}), and study its robustness .

\section{State of the art}\label{sec:sota}

In this section, 
we briefly discuss the state-of-the-art in fake detection methods, to set the stage for using those as a specific application domain for black-box-optimization (section \ref{sec:fd}).
Then, we introduce the general setup of the black-box optimization problem, and the set of algorithms that are used in the empirical comparison presented in this paper (section~\ref{sec:bbo}).

\subsection{Fake detectors}\label{sec:fd}
Due to the many AI-generated images on internet, it becomes important to be able to detect them.
DIRE~\citep{wang2023dire} is a method to detect images generated by Latent Diffusion Models (LDM)~\citep{rombach2022high}. 
It focuses on the error between an input image and its reconstruction counterpart using a pre-trained diffusion model. This method is based on the observation that the generated images can be approximately reconstructed by a diffusion model, while real images cannot. 
Several papers already mentioned artifacts making DIRE unreliable in a real-world scenario, in particular the impact of the image format on the prediction~\citep{clip}. 
Another line of work~\citep{coccomini2023detecting} explores the task using simpler, more traditional machine learning algorithms like Multi-Layer Perceptrons (MLPs), features extracted by Contrastive Language-Image Pretraining (CLIP, \citep{radford2021learning}), or traditional Convolutional Neural Networks (CNNs).
Generalization to unseen image generators is known as a critical issue for such supervised learning approaches: Universal Fake Detector~\citep{ojha2023fakedetect} uses a linear model trained on top of a generic feature extractor, for the sake of generalization and transfer, and GenDet~\citep{zhu2023gendet} uses an adversarial method aimed at solving the problem of unknown fake generators.
Another method that helps to detect LDM-generated images is the Local Intrinsic Dimensionality (LID) method \citep{lorenz2023detecting}. 
LID is employed to estimate the intrinsic dimensionality of a learned representation space, measuring the average distance between a point and its neighboring points. This approach is vital in characterizing the distinct properties of adversarial and natural samples in the latent space of a classifier. The multiLID method~\citep{lorenz2023detecting} extends this concept, combining locally discriminative information about growth rates in close proximity, which proves effective in detecting adversarial examples as well as images generated by diffusion models. 
Watermarking~\citep{cox2007digital} and on-device certification~\citep{nagm2021novel} are alternative methods for distinguishing between fake and real images.
Watermarking is designed to create imperceptible alterations to the images. 
These alterations, however, can be detected and decoded with the appropriate algorithms. 
The primary purpose of  watermarking is to ensure traceability of the images back to the model. 
Many of the current generation models already use built-in watermarking techniques. 

For instance, some latent diffusion models use Discrete Wavelet Transform together with Discrete Cosine Transform \citep{al2007combined}. 
While these transformations provide some protection from minor generative image alterations,  they are not resistant, for example, to image resizing. Furthermore, one can easily disable the watermarking.

More advanced watermarking methods, as proposed, for example, by \cite{fernandez2023stable}, merge watermarking into the generation process.
Watermarking methods are brittle to adversarial attacks. 
They include no-box attacks which do not need any knowledge about the method, e.g., using perceptual auto-encoders~\citep{fernandez2023stable}
or noising/denoising with a diffusion model~\citep{nie2022diffusion}; and white-box or black-box attacks (in particular, SA) which employs the watermark extractor or an API access to it~\citep{jiang2023evading}.
Fortunately, there are effective detectors for both no-box attacks and for SA.
We investigate how generic black-box optimization methods can be turned into adversarial attacks that are not detected by these detectors: therefore, we need detectors for this family of attacks.

\subsection{Black-box optimization}\label{sec:bbo}
An unconstrained optimization problem can be generally formulated as follows,
$\min f(x), \; x \in \Omega$.
where $\Omega$ denotes the search space, $f: \Omega \rightarrow \mathbb{R}^m$, and $m$ is the number of objectives. Note that we consider $m=1$ in this work, and without loss of generality, we assume a minimization problem. According to the domain of $\Omega$, we can consider problems as continuous optimization when $\Omega \subset \mathbb{R}^n$, as discrete optimization when $\Omega \subset \mathbb{Z}^n$, and pseudo-Boolean optimization (PBO) when $\Omega = \{0,1\}^n$, where $n$ is the dimensionality of the problem.
We deal with black-box optimization (BBO), in which algorithms can not obtain the exact definition of the objective function $f$ and constraint definitions regarding the structure of $f$. Evolutionary computation has been widely applied to solve BBO. For example, Evolution Strategies (ES), Differential Evolution (DE), etc., have achieved success in solving continuous BBO problems~\citep{BackKSWAKNVWY23}. 
The variants of these methods have also been applied for discrete BBO based on relaxation to a continuous problem~\citep{PanTL08,HamanoSNS22}. 
Moreover, EAs have been well-studied for pseudo-Boolean optimization (PBO)~\citep{DOERR2020106027}.
Some strategies have been commonly applied with specific  adjustments when solving different types of optimization problems. A recent study has inves\-tigated the performance of a discretized Covariance Matrix Adaptation Evolution Strategy 
(CMA-ES), addressing the adaptation of continuous optimization algorithms for the discrete domain \citep{ThomaserNVYBK23}. 

In this work, we work on continuous optimization by utilizing discrete optimization algorithms. 
Specifically, we investigate utilizing the techniques of the $(1+\lambda)$~EAs that are designed for PBO for continuous optimization~\citep{DOERR2020106027}.
The $(1+\lambda)$~EAs flip a number $\ell$ of variables to generate $\lambda$ new solutions from a single parent solution iteratively, and self-adaptive methods have been proposed to adjust the value of $\ell$ online, essentially detecting the optimal number of variables to be altered dynamically.
We consider various black-box optimization methods, which can all be found in \citep{nevergrad}. 
We include many algorithms, specified in Section \ref{algos}.
We selected, for readability, a sample of methods covering important baselines (such as random search), one representative method per group of related methods, and the overall best methods.
We refer to \citep{nevergrad} for the details about other algorithms.

\section{Methodology}\label{sec:met}
\subsection{\Lognormal{} mutations}

\label{sec:alg}
\label{sec:extensions}
We introduce in this section the general framework of the $(1+\lambda)$~EA with \lognormal{} mutation, which was proposed for the pseudo-Boolean optimization task, and illustrate a straightforward generalization of this algorithm to continuous and integer domains.

The \lognormal{} mutation was first described in \citep{Baeck1995,Baeck1996lognormal}, later refined in \citep{Kruisselbrink2011lognormal}. It stands out for its robust performance across various problem landscapes~\citep{DOERR2020106027}. 
Although it may not surpass other self-adjusting methods, such as the two-rate~\citep{DoerrGWY19} and the normalized bit mutation \citep{YeDB19}, which are good at converging to small mutation strength $\ell = 1$ for the classic theory-oriented problems OneMax, LeadingOnes, etc.,  \lognormal{} mutation consistently delivers competitive results for complex practical problems such as Ising models and Maximum Independent Vertex Set \citep{DOERR2020106027}.
\begin{algorithm2e}[t]\label{alg:lognormal}
\textbf{Input:} A given problem $f: \Omega \rightarrow \mathbb{R}$, where 
$n$ denotes the dimensionality of the problem, an initial value of the mutation rate $p \in (0,1)$, population size $\lambda > 0$, and a learning rate $\gamma = 0.22$\; %

\textbf{Initialization:} 
	Sample $x \in \Omega$ u.a.r.~and evaluate $f(x)$\; \label{alg:logNormalInit}

	\textbf{Optimization:}
	\For{$t=1,2,3,\ldots$}{
	    \For{$i=1,\ldots,\lambda$}{\label{alg:logNormalForBegin}
            $q \sim \mathcal{N}(0,1)$\;\label{alg:lognormalNormal}
	        $p^{(i)} = \left(1+\frac{1-p}{p}\cdot \exp(\gamma\cdot q)\right)^{-1}$  \; \label{alg:lognormalMutateP}
	        Sample $\ell^{(i)} \sim \Bin_{>0}(n,p^{(i)})$\;\label{alg:logNormalSample}
	        create $y^{(i)} \assign \mut(\ell^{(i)},x)$; evaluate $f(y^{(i)})$ \label{alg:logNormalY}
         \tcp*{See Alg.\ref{alg:mutation}}
	        }\label{alg:logNormalForEnd}%
         $i \assign \min\left\{ j \mid f(y^{(j)}) = \max\{f(y^{(k)}) \mid k \in [\lambda]\} \right\}$\;\label{alg:logNormalUpdateI}
         $p \assign p^{(i)}$\; \label{alg:logNormalUpdateP}
         $x^* \assign \arg\max\{f(y^{(1)}), \ldots, f(y^{(\lambda)})\}$ \label{alg:logNormalAssignX} \tcp*{ties broken by favoring the smallest index}
	\lIf{$f(x^*) \leq f(x)$}{$x \assign x^*$}\label{alg:logNormalUpdateX} 
 	}%
\textbf{Output:} $x, f(x)$
\caption{$(1+\lambda)$~EA$_{log-n}$ with \lognormal{} mutation.
\label{alg:log-n}
}
\end{algorithm2e}
Following the setup in recent work \citep{DOERR2020106027} on pseudo-Boolean optimization, we provide a concise formulation of a $(1+\lambda)$~EA with the \lognormal{} mutation in Algorithm~\ref{alg:lognormal}.
The algorithm starts from a randomly initialized point (line \ref{alg:logNormalInit}) and generates $\lambda$ offspring by the \textsc{Mutate} function in the for-loop (lines \ref{alg:logNormalForBegin}-\ref{alg:logNormalForEnd}).
We denote the mutation strength $\ell$ as the number of variables to be altered, and this number is generated for each newly generated solution candidate, i.e., offspring, by first mutating the mutation rate $p$. In practice, $p$ is mutated according to a \lognormal{} distribution (line \ref{alg:lognormalMutateP}, where $\mathcal{N}(0,1)$ is a normally distributed random variable with expectation zero and standard deviation one, from which $q$ is sampled in line \ref{alg:lognormalNormal}),
and this rule maintains that the median of the distribution of new mutation rates $p^{(i)}$ is equal to the current mutation rate $p$. 
$p^{(i)}$ then determines the value of $l^{(i)}$ (line \ref{alg:logNormalSample}), and the corresponding offspring is created and evaluated in line \ref{alg:logNormalY}. 
Specifically, $\ell^{(i)}$ is sampled from a binomial distribution Bin$_{>0}(n,p^{(i)})$ (line \ref{alg:logNormalSample}), where $n$ is the dimensionality, 
and the sampling is repeated until obtaining $\ell^{(i)} > 0$.
In case of ties in the objective function values, the mutation rate used for the first best of the newly generated solutions (line \ref{alg:logNormalUpdateI}) is taken for the next iteration (line \ref{alg:logNormalUpdateP}). 
The best solution $x^*$ is updated in line \ref{alg:logNormalUpdateX} if the best offspring is equal to or better than the current parent solution.
 
\subsection{Modifications of the original \lognormal{} algorithm}
\label{sec:alg:extend}

\begin{algorithm2e}[tb]
\textbf{Input:} a solution $x \in \varprod_{i=1}^n \mathcal{X}_i$, and the mutation strength $\ell \in \{1,2, \ldots, n\}$, where $n$ is the dimensionality of the problem\;
Sample $\ell$ pairwise different positions $i_1,\ldots,i_\ell \in [n]$ u.a.r.\;
$y \leftarrow x$\;
\For{$j=1,\ldots,\ell$}{
    \Repeat{$k \neq x_{i_j}$}{$k \sim \mathcal{U}(\mathcal{X}_{i_j})$
    \tcp*{Sample $k$ u.a.r.~from domain $\mathcal{X}_{i_j}$}}
    $y_{i_j} \leftarrow k$\;
}
 \textbf{Output:} $y$\;
\caption{$\mut(\ell, x)$}
\label{alg:mutation}
\end{algorithm2e}
 
Note that in Algorithm~\ref{alg:lognormal} we do not specify the domain $\Omega$ of $x$.
The \lognormal{} mutation has been commonly applied for the pseudo-Boolean optimization, i.e., $\Omega = \{0,1\}^n$, and the $\mut$-operator flips $l^{(i)}$ bits that are selected uniformly at random.
In this work, we introduce a generalization of the $\mut$-operator, as presented in Algorithm~\ref{alg:mutation}, to extend the \lognormal{} mutation to other domains. This generalized $\mut$
is applicable for arbitrary search domains $\Omega = \varprod_{i=1}^n \mathcal{X}_i$, where $\mathcal{X}_i \in \{\{0,1\}, \mathbb{Z}, \mathbb{R}, \mathbb{N}\}$.
It samples a new value that is distinct from the current one at random  %
from the respective domain for each of $\ell$ variables, which are selected (uniformly at random) for mutation.
This technique has also been adopted by Nevergrad \citep{nevergrad}, which automatically adapts algorithms designed for continuous domains to also work for discrete domains and vice versa.

\section{Experimental results on the Nevergrad Black-box optimization benchmarks }\label{lognormcompa}\label{sec:experiments}

\subsection{Parameter settings \& Ablation}

While the \lognormal{} mutation can control mutation rates of the $(1+\lambda)$~EA online, Algorithm~\ref{alg:log-n} still comprises three hyperparameters, i.e., the initial value of $p$, population size $\lambda$, and $\gamma$, that can affect the algorithm's performance. We set $\gamma = 0.22$ following the suggestion in previous studies \citep{Kruisselbrink2011lognormal,DOERR2020106027}. For the other two hyperparameters, we test two values $0.2$ and $0.8$ for $p$ and various population sizes $\lambda$. The detailed combinations are presented in Table~\ref{lognv}.

We denote \lognormal{} as the standard setting. BigLognormal experiments are performed with an increase in population size (respectively, large increase for HugeLognormal, and decrease for SmallLognormal).
XLognormal obtains a greater initial mutation rate.
OLN (Optimistic Lognormal) experiments the combination with Optimism as performed in Nevergrad for making deterministic algorithms compatible with noisy optimization (see \cref{optmodifiers}).

\begin{table}[t]
\begin{center}
\begin{tabular}{ccc|ccc}
\toprule
Name & Initial $p$ & $\lambda$  & Name & Initial $p$ & $\lambda$  \\
\midrule
Lognormal (Standard setting) & 0.2 & 12  & SmallLognormal & 0.2 & 4 \\
BigLognormal & 0.2 & 120 & XLognormal & 0.8 & 12 \\
HugeLognormal & 0.2 & 1200 & XSmallLognormal & 0.8 & 4  \\ 
OLN(Combined with bandits) & 0.2 & 12 &&& \\
\bottomrule
\end{tabular}
\end{center}
\caption{\label{lognv} Variants of \lognormal{} mutations used in the experiments. OLN is a combination with bandits, for managing reevaluations in a noisy optimization context.}
\end{table}

We compare the settings of Algorithm~\ref{alg:log-n} with several algorithms, such as random search, CMA, and the anisotropic adaptive algorithm~\citep{anisotropic1,anisotropic2} provided by Nevergrad. \lognormal{} with the standard parameterization from~\citep{Baeck1996lognormal} performs essentially well across the 14 tested problems (see Appendix~\ref{appendix:compare-log-variants}), though the ``Big'' variant is an interesting outsider.
We note that the default parametrization of \lognormal{} fails mainly on topology optimization (for which anisotropic methods perform great) and on noisy problems (which are irrelevant for our context of attacking deterministic fake detectors, and we note that a good solution in that setting is to use a combination with optimism in front of uncertainty as proposed in \citep{nevergrad}). We therefore keep the standard \lognormal{} and some ``Big'' counterparts for our application to fake detectors.

\subsection{Robust performance of \lognormal}\label{sec:nev}

The \lognormal{} algorithm has already shown competitive results in existing benchmark studies~\citep{DoerrGWY19}, and in the present paper we examine its performance by comparing it on extensive benchmarks provided by Nevergrad.
Table~\ref{div} shows the diversity of the benchmarks in Nevergrad, where each benchmark is accompanied with a list of baselines.

Since NGOpt is a ``wizard'' provided by Nevergrad that is tuned for selecting automatically a proper algorithm for each problem, we use it as a baseline for comparison.
We run \lognormal{} and NGOpt and proposed default methods.

Tables~\ref{dag} and \ref{dag2} list the rank of the \lognormal{} algorithm and of NGOpt:
Table~\ref{dag} lists the problems in which the benchmark outperforms NGOpt, and Table~\ref{dag2} presents the results of the other benchmarks.

We observe that the \lognormal{} algorithm can outperform NGOpt on 17 out of 41 benchmarks and perform better than $50\%$ of all algorithms on 29 benchmarks. Recall that Nevergrad contains a diverse set of benchmarks, and it is well-known that we can not expect one algorithm to perform the best for all problems. The lognormal algorithm shows robust and competitive performance for Nevergrad benchmarks.

\begin{figure}[t]\centering
\includegraphics[width=.95\textwidth]{{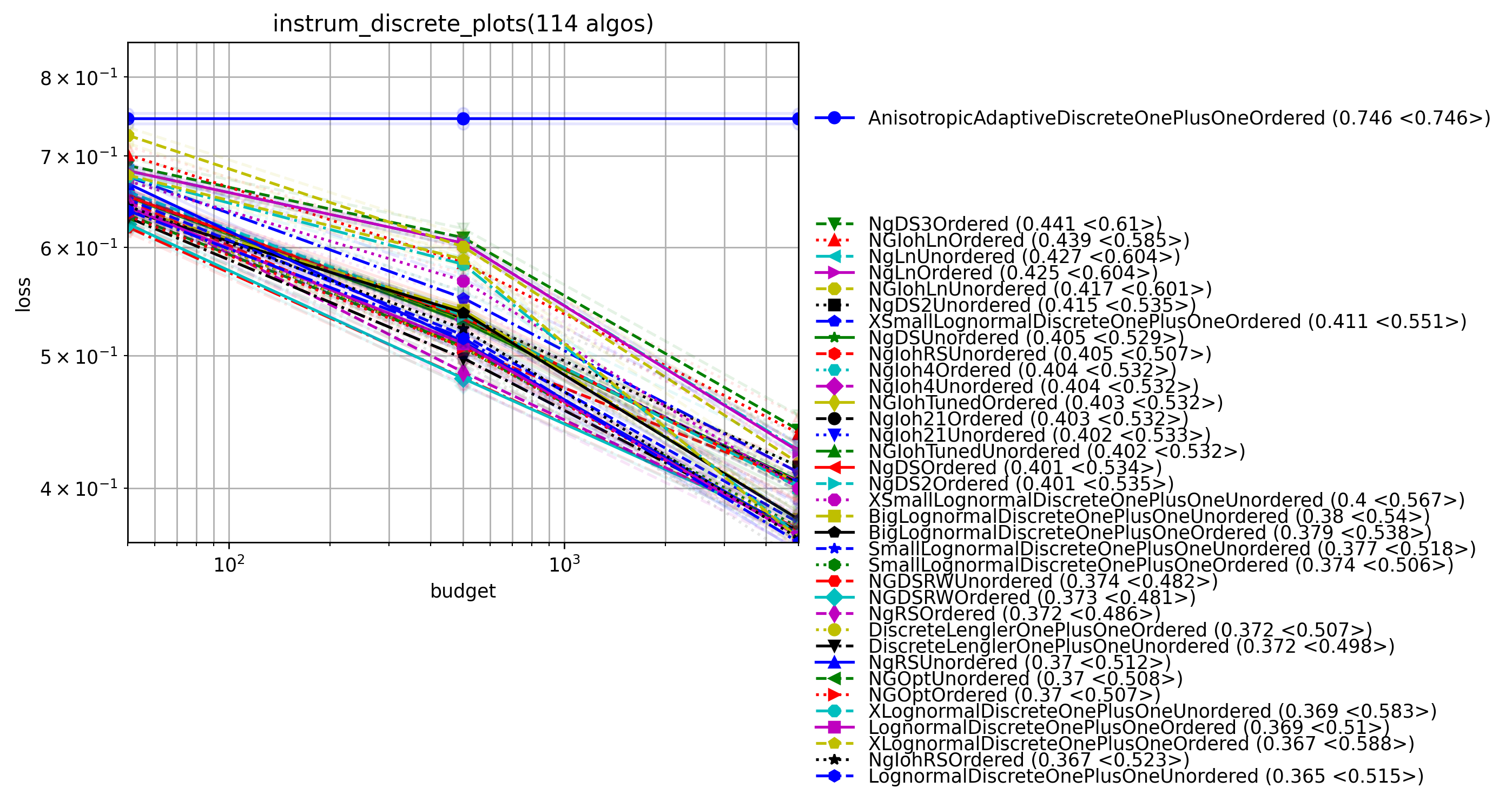}}
\includegraphics[width=.95\textwidth]{{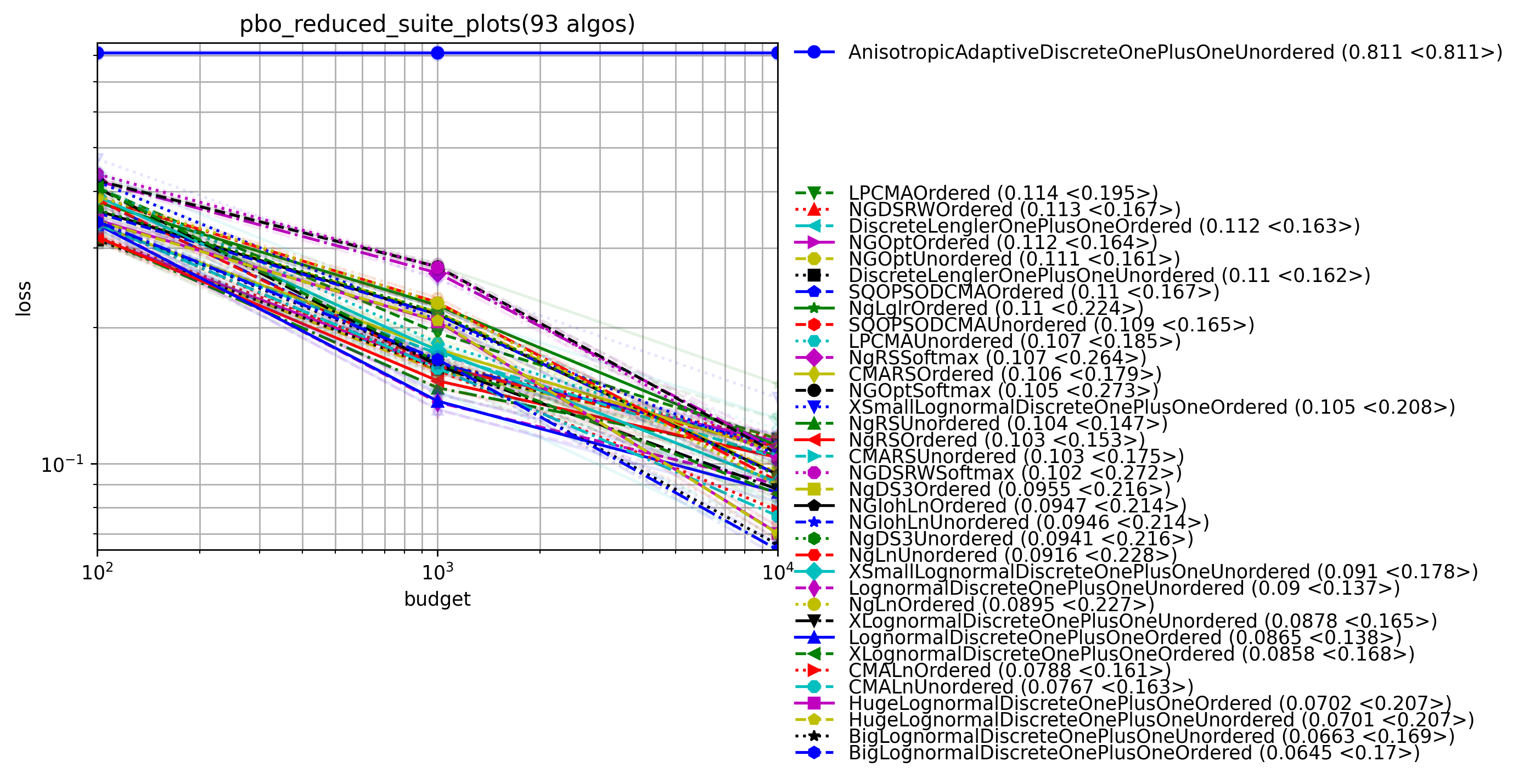}}
\caption{\label{disc}Results on instrum-discrete (top, 35 best methods and the worst method out of 98 methods run on this benchmark), pbo-reduced (bottom, 35 best and the worst method out of 156 methods run on this benchmark). \lognormal{} is simple but good. CMALn is a combination of CMA and \lognormal{} (used as a warmup during the early 10\% of the budget): on PBO all strong methods use \lognormal{} at some point. }
\end{figure}

Since the \lognormal{} algorithm was originally proposed for discrete optimization, we first present the detailed results of \lognormal{} variants for the two discrete benchmarks in Fig.~\ref{disc}. In Fig.~\ref{disc}, the y-axis represents the loss of algorithms for tested budgets (presented on the x-axis), and algorithm labels are annotated by their average loss for all the maximum budget between parentheses and (for checking stability) the average loss for the tested budgets excluding the largest one between brackets (if there are at least three budget values). For readability, we present only the 35 best results (and the worst, for scale) for each benchmark. We can observe that the \lognormal{} variants obtain the best performance for the two discrete benchmarks.

\begin{table}[t]\centering
\small
\begin{tabularx}{\linewidth}{p{.35\textwidth}XXX}
\toprule
\multicolumn{4}{c}{Cases in which \lognormal{} outperforms NGOpt}\\
\midrule
Problem & Rank of \lognormal{} & Num algorithms & Rank of NGOpt \\
\midrule
deceptive**  & 0 & 61&  6 \\
fishing**  & 0 & 63&  56 \\
multiobj-example-many-hd  & 1 & 67&  48 \\
yatuningbbob  & 4 & 93&  54 \\
multiobj-example-hd  & 5 & 69&  55 \\
multiobj-example  & 9 & 63&  29 \\
yaonepensmallbbob  & 9 & 79&  15 \\
nano-seq-mltuning  & 10 & 23&  11 \\
yasmallbbob  & 10 & 96&  19 \\
nano-naive-seq-mltuning  & 11 & 23&  14 \\
zp-pbbob  & 13 & 41&  26 \\
nano-veryseq-mltuning  & 14 & 26&  16 \\
pbo-reduced-suite*  & 16   & 156& 20 \\
verysmall-photonics**  & 26 & 64&  30 \\
verysmall-photonics2**  & 33 & 67&  37 \\
pbbob  & 35 & 63&  42 \\
ultrasmall-photonics2**  & 44 & 83&  72 \\
\bottomrule
\end{tabularx}
\caption{\label{dag}Results on benchmarks from the Nevergard benchmarking suite for which \lognormal{} outperforms NGOpt. The rank, computed as described in Section \ref{ranking}, is between $0$ and $num-algos - 1$. We compare \lognormal{} to NGOpt. Cases in which \lognormal{} is outperformed by NGOpt are in Table \ref{dag2}. We note that  results on continuous problems are not bad, in particular for hard problems: Deceptive (which is designed for being hard), some multi-objective benchmarks (in particular the many-objective case), PBBOB which uses difficult distributions of random translations for optima, very low budget problems such as YaTuningBBOB, and difficult low budget Photonics or Fishing problems. * denotes discrete problems, ** denotes single-objective problems which are highly multimodal and difficult.}
\end{table}

\subsection{Competitive results for deceptive problems}\label{selected}

\begin{figure}[t]\centering
\includegraphics[width=.95\textwidth]{{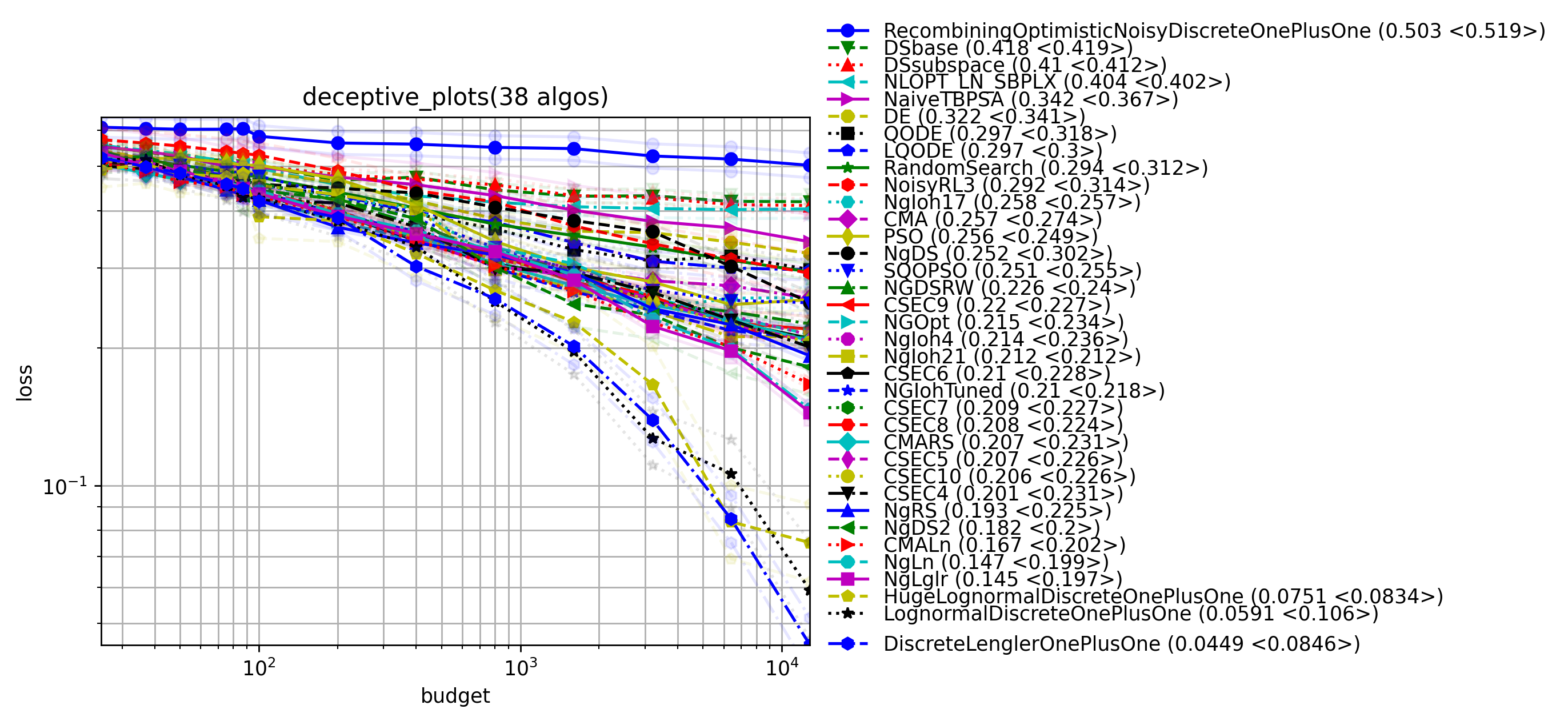}}
\caption{\label{deceptive}Results on the Deceptive benchmark in Nevergrad. X-axis: budget. Y-axis:  average normalized (linearly to $[0,1]$, for each benchmark) loss. We observe that CMALn (CMA with \lognormal{} warmup) outperforms CMARS (CMA with random search warmup), which outperforms CMA, on this hard benchmark. The best algorithms are based on \lognormal: CMALn (resp. NgLn) uses CMA (resp. NGOpt) for local optimization after \lognormal. Lengler also performs well, showing that discrete algorithms can be competitive for continuous problems, in the hardest cases, as a warmup or as a standalone method. The CSEC codes are all variants of NgIohTuned: they are good, but still outperformed by codes based on Lognormal. We note an excellent performance of the Lengler method adapted to continuous problems, in particular for the greatest values of the budget, though Table \ref{dag} shows that \lognormal{} was better over the different budget values $(25, 37, 50, 75, 87, 100, 200, 400, 800, 1600, 3200, 6400, 12800)$ for the criterion defined in Section \ref{ranking}.} 
\end{figure}

Recall that in the ranks in Table~\ref{dag}, the \lognormal{} algorithm performs best for the ``deceptive'' and ``fishing'' benchmarks and outperforms NGOpt on many difficult multimodal benchmarks (marked with **). %
We plot in Fig.~\ref{deceptive} the detailed results of the best of 61 algorithms on the deceptive benchmark. The deceptive benchmark combines many random translations of hard problems in many dimensionalities, including (i) a problem with a long path to the optimum, which becomes thinner and thinner close to the optimum, (ii) a problem with infinitely many local minima, (iii) a problem with condition number growing to infinity as we get closer to the optimum.
We note that on this hard continuous benchmark, the \lognormal{} algorithm performs well across all the tested budgets.
Therefore, due to its robust performance across a number of benchmarks and particularly competitive performance for the hardest benchmarks, we utilize the \lognormal{} mutation for the following fake detector scenarios.  

\section{Stress test for fake detectors}\label{sec:fake}

We consider stress tests for fake detectors. More precisely, we add imperceptible noise $e$ to the image $x$, with this noise chosen so that fake detectors fail: typically $x$ is a fake image detected as fake by a detector $\cal D$ and we find a small $e$ such that $x+e$ is classified as non-fake by ${\cal{D}}$.
Black-box algorithms can be straightforwardly applied to attacking a fake detector ${\cal{D}}$ (which returns the estimated probability ${\cal{D}}(x)$ that an image $x$ is fake) by defining $loss(e)={\cal{D}}(e+x)$ and minimizing $loss$ on a domain $D$ (typically $[-0.03,0.03]^t$, where $t$ is the shape of the image tensor\footnote{
In this case, we assume that pixels are in the range $[0,1]$.
Due to maximum values in image representations we might clamp the values of $e+x$.}) using a black-box algorithm. 
Our attacks (Section \ref{sec:dire}) consider the fake detector as a black-box, so we do not have gradients. 
Then, in Section \ref{defense}, we create defense mechanisms by detecting various black-box attacks, showing that the \lognormal{} attack is not detected by a detector created for a classical attack (such as Square Attack) and, therefore, needs new ad hoc detectors.

\subsection{Black-box attacking fake detectors}\label{sec:dire}\label{perf}

The fake detector we consider is the universal fake detector~\citep{ojha2023fakedetect}.
Precisely, we consider the datasets as described in Table \ref{datasets}. IN500- and IN500+ refer to real images, \G{}, \PG{}, \BG{} and \LDM{}-200 refer to images created by image generators. In the present paper, Clean means unattacked, neither by no-box nor by black-box attacks.

\begin{table}[t]
\renewcommand{\arraystretch}{1.5} %
\begin{tabular}{p{2.6cm}p{13.0cm}}
\toprule
Name & Description \\
\midrule
Dataset1 &        2k images from \PG{}, 2k images from \BG{} and 1k images from \LDM{}-200 for a total of 5k fake images. Used to benchmark variations of \lognormal{} attacks and square attack. \\
\hline
Dataset2-DP & Same as Dataset1, plus 5k real images from \IN{}500- and their purified versions by DiffPure with parameter 0.1\\
\hline
Dataset2-IR & Same as Dataset1, plus 5k real images from \IN{}500- and their purified versions by ImageRephrase with parameter 0.1\\
\hline
 Dataset2-SA       & Same as Dataset1, plus 5k real images from IN500- and their attacks by SA with budget=10k and $l^\infty=0.01$. Total = 20k images. \\
\hline
Dataset2-LN       & Same as Dataset1, plus 5k real images from IN500- and their attacks by \lognormal{} (algo1) with budget=10k and $l^\infty=0.01$. Total = 20k images. \\
\hline
Dataset3 &         1k real images from \IN{}500+ + 500 fake images from \PG{} of different classes than Dataset2 + 500 fake images from \G{} and their attacked counterpart. Total $=$ 4k images.  \\
\hline
Dataset3-IR0.1, IR0.2, IR0.3 & Purified versions of Dataset3 based on ImageRephrase, with parameter 0.1, 0.2, 0.3. \\
\hline
Dataset3-DP0.1, DP0.2, DP0.3 & Purified versions of Dataset3 based on DiffPure, with parameter 0.1, 0.2, 0.3. \\
\hline
Dataset3-SA ($L$,$B$)   & Real images from Dataset3 + fake images from Dataset3 attacked by SA with various amplitudes ($L\in \{ 0.01, 0.03, 0.05\}$) and budget ($B\in\{1000, 10000 \}$). \\
\hline
Dataset3-LN ($L$,$B$)   & Real images from Dataset3 +  fake images from Dataset3 attacked by LN (algo1 to algo5) with various $L\in \{ 0.01, 0.03, 0.05\}$ and $B\in\{1000, 10000 \}$. \\
\hline
Dataset4    & 1k fake images from Dataset3, attacked with \lognormal{} attacks (algo1).\\
\bottomrule
\end{tabular}
\caption{Datasets used in the study. Dataset3 will be used as a test set for the detectors trained on Dataset2 and Dataset4, will be used for testing the transfer of detectors trained on SA attacks to \lognormal{} attacks. }
\label{datasets}
\end{table}

{\bf{Critical cases.}} We consider the detection of attacks (no-box or black-box attacks) aimed at evading fake detectors. 
So, we are particularly interested in correctly detecting attacks in two cases (1) fake images attacked for making them appear genuine and (2) real, unaltered images (clean). 
The other cases are less critical: 
(1) the case of real images attacked so that they will look fake: this will usually not prevent the original images from being classified as real, making the situation less annoying. 
(2) the case of clean fake images (erroneously viewing them as attacked will not prevent a fake detector from working).
Therefore, besides our accuracies on datasets, we will present accuracies on the critical part of these datasets.

\begin{figure}
    \begin{tabular}{cccc}
        \includegraphics[width=0.22\textwidth]{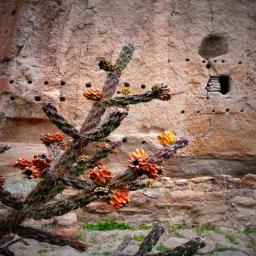} &
        \includegraphics[width=0.22\textwidth]{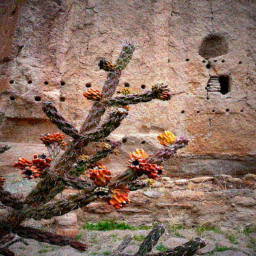} &
        \includegraphics[width=0.22\textwidth]{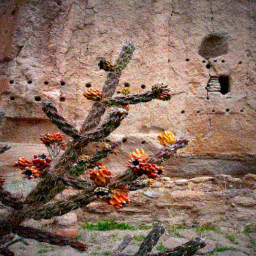} &
        \includegraphics[width=0.22\textwidth]{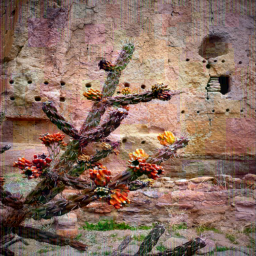} \\
        \includegraphics[width=0.22\textwidth]{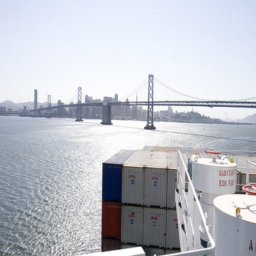} &
        \includegraphics[width=0.22\textwidth]{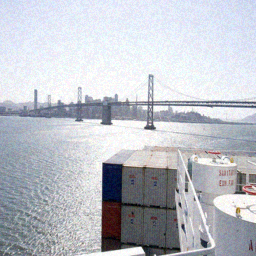} &
        \includegraphics[width=0.22\textwidth]{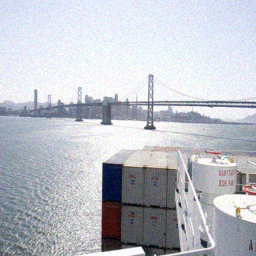} &
        \includegraphics[width=0.22\textwidth]{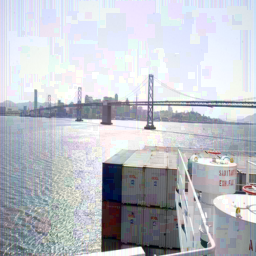} \\
        \parbox{0.22\textwidth}{\centering Original} &
        \parbox{0.22\textwidth}{\centering algo 1} &
        \parbox{0.22\textwidth}{\centering algo 5} &
        \parbox{0.22\textwidth}{\centering SA}
    \end{tabular}
    \label{fig:example_images}
    \caption{Example of attacked images. All attacks are done with a budget of 10k queries and $l^\infty=0.03$.}
\end{figure}

We first attack the fake detector with the classical SA, and then with generic black-box optimization methods from \cref{sec:bbo}.
We give a brief overview here and full details are in \cref{smooth}.
Nevergrad provides modifiers dedicated to domains shaped as images. These modifiers can be applied to any black-box optimization algorithm. For example, the prefix \emph{Smooth} means that the algorithm periodically tries to smooth its attack $x$, if the loss of $Smooth(x)$ is better than the loss of $x$, then $x$ is replaced by $Smooth(x)$ in the optimization algorithm.
We add two additional modifiers specific from adversarial attacks. Both are inspired from SA, though our modifications have, by definition, less horizontal and vertical artifacts.
First, G (Great) means that we replace $loss(x)$ by $loss(0.03\times \text{sign}(x))$ when the allowed norm of the attack is $0.03$ for the $l^\infty$ norm. Second, SM (smooth) means that we replace the loss function $f$ by $f(convolve(x))$, where $convolve$ applies a normal blurring with standard deviation $3/8$. GSM means that we apply both G and SM.
For example, we get GSM-SuperSmoothLognormalDiscreteOnePlusOne by applying G, SM and SuperSmooth as modifiers on top of the standard \lognormal{} method, and various methods as in Tab \ref{tab:algorithms} (top). 
Results are presented in Tab. \ref{attackfake} (bottom). Examples of attacked images are presented in Figure \ref{fig:example_images}.
Basically, when we have a black-box access to a fake detector, we can attack it either by SA or by \lognormal{} with a budget of 10k queries and $l^\infty=0.03$.

\begin{table}[t]
\centering
\renewcommand{\arraystretch}{1.2}
\resizebox{0.99\linewidth}{!}{
    \begin{tabular}{cc|cc}
    \toprule
    Algorithms & Alias & Algorithms & Alias \\
    \midrule
    GSM-SuperSmoothLognormalDiscreteOnePlusOne & algo1 & LognormalDiscreteOnePlusOne & algo4 \\
    G-SuperSmoothLognormalDiscreteOnePlusOne & algo2 & GSM-BigLognormalDiscreteOnePlusOne & algo5 \\
    SuperSmoothLognormalDiscreteOnePlusOne & algo3 & G-BigLognormalDiscreteOnePlusOne & algo6 \\
    \bottomrule
    \end{tabular}
}
\vspace*{0.5\baselineskip}
\begin{tabular}{c|*{7}{c}}
\toprule
Algorithm & SA & algo1 & algo2 & algo3 & algo4 & algo5 & algo6 \\
\midrule
Success rate on dataset1       & 100\%   & 91.0\% & 94.1\% & 64.6\% & 82.0\% & 99.2\% & 99.1\% \\
\bottomrule
\end{tabular}
\caption{\label{tab:algorithms}Top: Variations of the \lognormal{} algorithm considered. \label{attackfake}Bottom: Success rate when attacking the fake detector. The success rate is computed solely on attacks of correctly classified clean images. All attacks have 10k budget and $l^\infty=0.03$. }
\end{table}
\subsection{From attack to defense: detecting attacks}\label{defense}
 A recent trend is the denoising of watermarked images for evading the detection: whereas the watermark makes it possible to detect that an image is fake, the denoised version goes undetected. 
 However, a defense is to detect such a denoising (e.g.~by DiffPure or ImageRephrase), or other forms of attacks (such as SA). That way, our work both provides (i) new attacks and (ii) detectors for those attacks, to be used as an additional step (detection of evasion) in fake detectors.

\subsubsection{No-box attacks (purifiers) and their detection}

While watermarking techniques can be made robust against some image alterations such as resizing or JPEG compression as shown by~\cite{fernandez2023stable}, they remain vulnerable to no-box purification attacks~\citep{saberi2023robustness} that destroy the hidden message. Additionally, the distribution change introduced by no-box purification can impact negatively the performance of a fake detector, as shown in Table \ref{tab:no_box_ufd}. We will next show that one can easily detect these attacks if we have access to the purification method.

We consider two purifiers used by \cite{noboxattack}:
DiffPure based on guided diffusion and ImageRephrase based on latent diffusion. We consider the strength/steps parameter in the set \{0.1, 0.2, 0.3\}.
These two purifiers lead to an average PSNR (Peak Signal-to-Noise Ratio) between 20 and 30, which is a noticeable yet acceptable quality loss.
As expected, the PSNR of the distortion increases with the strength parameter.

\begin{table}[t]
\centering
\begin{tabular}{*{8}{c}}
\toprule
 & \multicolumn{7}{c}{Attacks} \\
\cmidrule(lr){2-8}
& Clean & DP 0.1 & DP 0.2 & DP 0.3 & IR 0.1 & IR 0.2 & IR 0.3 \\
\midrule
FNR & 5.2\%   & 23.42\% & 34.4\% & 40.1\% & 11.3\% & 16.8\% & 26.2\% \\
\midrule
PSNR  & NA & 27.9 & 24.4 & 22.1 & 26.6 & 24.4 & 22.7 \\
\bottomrule
\end{tabular}
\caption{\label{tab:no_box_ufd} The second row shows the false negative rate (FNR) at a threshold of 0.5 for the universal fake detector on clean Dataset1 consisting of only fake images and different purified versions of this dataset. DP stands for DiffPure while IR stands for ImageRephrase. 
The third row corresponds to the average PSNR.}
\end{table}

For detecting the no-box attacks, we train a classifier, specified in \cref{tab:learner}, on Dataset2-DP or Dataset2-IR. We experiement with both ResNet50 and SRnet~\citep{boroumand2018deep} and the latter performs better overall. We use Dataset3 as a test set (so that the distributions of images differ) with images attacked by same purifier but with different parameters in $\{0.1, 0.2, 0.3\}$. 
The results of the detection of latent purifiers and guided diffusion purifiers are shown in Table \ref{tab:no_box_results_latent} and Table \ref{tab:no_box_results_dp}. We achieve less than 5\% FPR and FNR across the hold-out training dataset and the critical part of testset, although the FPR is relatively higher for the full testset that can be explained by having a sort of transfer from detecting DP or IR to detecting images generated by G.

In short, training on a purifier with a specific parameter allows us to detect that same purifier with different parameters and for different image distributions.

\subsubsection{The detection of black-box attacks}\label{sa_detection}
For detecting square attacks, we train a deep net using a similar setup, as the SRnet model still performs better than ResNet in this scenario. We also employ data augmentation techniques: horizontal flip, random crop and color jitter which empirically makes the model more robust to different attack parameters, while allowing us to train using only one attack parameter budget=10k and $l^\infty=0.01$.
The training is done on Dataset2-SA. For testing, we use Dataset3-SA using different parameters than those used during training. We then test the transfer of the SA detector on Dataset4, corresponding to \Lognormal{} attacks. 

Table \ref{tab:sa_results} summarizes the results of the SA detector, for detecting SA and as a detector of \lognormal{} 
 attacks.
We observe that the transfer to detecting \lognormal{} attacks is very poor. So, we need to include such attacks in our training for improving the defense.

{\bf{A new detector for \lognormal{} attacks.}}
\label{ln_detector}
We have seen that detectors of SA do not detect our \lognormal{} attacks.
For detecting \lognormal{} attacks, we use the same setup as in section \ref{sa_detection} for creating a new detector. Dataset2-LN is used during training with the attacked images now obtained with GSM-SuperSmoothLognormalDiscreteOnePlusOne budget=10k and $l^\infty=0.01$ and again, the dataset is split to 80\% train, 10\% test, 10\% validation.
For testing, we use Dataset3 with the images attacked using different variations of \lognormal{} and parameters (Budget and $l^\infty$). Table \ref{tab:lognormal_results} presents the results of the \lognormal{} detector: we observe that the learning was made on images attacked by algo1 only and we get positive results for all \lognormal{} variants.

\begin{table}[t]
    \centering
    \begin{tabular}{ccccc}
    \toprule
\rowcolor{lightgray!30}
         \multicolumn{2}{c}{Dataset} & FPR$\downarrow$ & FNR$\downarrow$ & AUC$\uparrow$ \\
    \midrule
         (same dist) & Dataset2(param=IR0.1) hold-out & 0.9\% & 0.3\% & 0.99 \\
    \midrule
\rowcolor{gray!30}    \multicolumn{5}{|c|}{Dataset 3, full (critical and non critical images)}\\
\hline
\rowcolor{lightgray!30}
         \multirow{3}{*}{(different dist)} & Dataset3 (param=IR0.1) & 23.1\% & 0.1\% & 0.96 \\
         & Dataset3 (param=IR0.2) & 23.1\% & 0.2\% & 0.96 \\
\rowcolor{lightgray!30}
         & Dataset3 (param=IR0.3) & 23.1\% & 0.2\% & 0.95 \\
         \rowcolor{gray!30} 
    \multicolumn{5}{|c|}{Dataset 3, critical part}\\
\hline
    \rowcolor{lightgray!30}
         \multirow{3}{*}{(different dist)} & Dataset3 (param=IR0.1) & 1.7\% & 0.0\%  & 0.99 \\
         & Dataset3 (param=IR0.2) & 1.7\% & 0.0\% &  0.99 \\
\rowcolor{lightgray!30}
         & Dataset3 (param=IR0.3) & 1.7\% & 0.0\% & 0.99 \\     
    \bottomrule
    \end{tabular}
    \caption{False positive and false negative rates for purified images detection at a threshold of (0.5) alongside with the AUC score. The purified training images correspond to the latent-purifier (class ``ImageRephrase'') with parameter 0.1. We observe good results, in particular for critical cases.
    }
    \label{tab:no_box_results_latent}
\end{table}

\begin{table}[t]
    \centering
    \begin{tabular}{ccccc}
    \toprule
\rowcolor{lightgray!30}
         \multicolumn{2}{c}{Dataset} & FPR$\downarrow$ & FNR$\downarrow$ & AUC$\uparrow$ \\
    \midrule
         (same dist) & Dataset2(param=DP0.1) hold-out & 1.9\% & 3.9\% & 0.99 \\
    \midrule
    \rowcolor{gray!30}    \multicolumn{5}{|c|}{Dataset 3, full (critical and non critical)}\\
\hline
\rowcolor{lightgray!30}
         \multirow{3}{*}{(different dist)} & Dataset3 (param=DP0.1) & 23.0\% & 3.7\% & 0.88 \\
         & Dataset3 (param=DP0.2) & 23.0\% & 4.0\% & 0.87 \\
\rowcolor{lightgray!30}
         & Dataset3 (param=DP0.3) & 23.0\% & 4.2\% & 0.87 \\

         \rowcolor{gray!30} 
    \multicolumn{5}{|c|}{Dataset 3, critical part}\\
\hline
    \rowcolor{lightgray!30}
         \multirow{3}{*}{(different dist)} & Dataset3 (param=DP0.1) & 2.2\% & 3.0\% & 0.99\\
         & Dataset3 (param=DP0.2) & 2.2\% & 4.6\%  & 0.99 \\
\rowcolor{lightgray!30}
         & Dataset3 (param=DP0.3) & 2.2\% & 5.4\% & 0.99 \\     
    \bottomrule
    \end{tabular}
    \caption{False positive and false negative rates for purified images detection at a threshold of (0.5) alongside with the AUC score. The purified training images correspond to the guided-diffusion purifier (class ``DiffPure'') with parameter 0.1. We observe good results, in particular when we restrict the analysis to critical cases (see the specification of critical cases in \cref{sec:dire}.}
    \label{tab:no_box_results_dp}
\end{table}

\begin{table}[t]
    \centering
    \begin{tabular}{ccccc}
    \toprule
\rowcolor{lightgray!30}
         \multicolumn{2}{c}{Dataset} & FPR$\downarrow$ & FNR$\downarrow$ & AUC$\uparrow$ \\
    \midrule
         (same dist) & Dataset2-SA hold-out & 1.8\% & 0.5\% & 0.99 \\
    \midrule
         \multirow{6}{*}{(different dist)} & \cellcolor{lightgray!30}Dataset3-SA (B=10k, L=0.01) & \cellcolor{lightgray!30} 0.8\% & \cellcolor{lightgray!30} 0.9\% & \cellcolor{lightgray!30} 0.99 \\
         & Dataset3-SA (B=10k, L=0.03)& 0.8\% & 0.0\% & 0.99 \\
         & \cellcolor{lightgray!30}Dataset3-SA (B=10k, L=0.05) & \cellcolor{lightgray!30} 0.8\% & \cellcolor{lightgray!30} 0.0\% & \cellcolor{lightgray!30} 0.99 \\
         & Dataset3-SA (B=1k, L=0.01) & 0.8\% & 0.0\% & 0.99 \\
         & \cellcolor{lightgray!30}Dataset3-SA (B=1k, L=0.03) & 
         \cellcolor{lightgray!30} 0.8\% & \cellcolor{lightgray!30} 0.0\% & \cellcolor{lightgray!30} 0.99 \\
         & Dataset3-SA (B=1k, L=0.05) & 0.8\% & 0.0\% & 0.99 \\
    \midrule
            (transfer to \lognormal) & \cellcolor{lightgray!30}Dataset4 & \cellcolor{lightgray!30} NA & \cellcolor{lightgray!30} 79.5\% & \cellcolor{lightgray!30} NA \\
    \bottomrule
    \end{tabular}
    \caption{False positive and false negative rates for square attack detection at a threshold of (0.5) alongside with the AUC score. B stands for the budget used for the attack and L stands for the accepted $l^\infty$ distance. Last row: transfer to \lognormal{} detection. }%
    \label{tab:sa_results}
\end{table}

\begin{table}
    \centering
    \begin{tabular}{ccccc}
    \toprule
\rowcolor{lightgray!30}
         \multicolumn{2}{c}{Dataset} & FPR$\downarrow$ & FNR$\downarrow$ & AUC$\uparrow$ \\
    \midrule
         (same dist) & Dataset2-LN hold-out & 0.7\% & 2.1\% & 0.99 \\
    \midrule
         \multirow{7}{*}{(different dist)} & \cellcolor{lightgray!30} Dataset3-LN (algo1, B=10k, L=0.01) & \cellcolor{lightgray!30} 4.1\% & \cellcolor{lightgray!30} 5.4\% & \cellcolor{lightgray!30} 0.98 \\
         & Dataset3-LN (algo1, B=10k, L=0.03) & 4.1\% & 2.3\% & 0.99 \\
         & \cellcolor{lightgray!30} Dataset3-LN (algo1, B=10k, L=0.05) & \cellcolor{lightgray!30} 4.1\% & \cellcolor{lightgray!30} 2.9\% & \cellcolor{lightgray!30} 0.99 \\
         & Dataset3-LN (algo2, B=10k, L=0.03) & 4.1\% & 4.1\% & 0.98 \\
         & \cellcolor{lightgray!30} Dataset3-LN (algo3, B=10k, L=0.03) & \cellcolor{lightgray!30} 4.1\% & \cellcolor{lightgray!30} 11.7\% & \cellcolor{lightgray!30} 0.95 \\
         & Dataset3-LN (algo4, B=10k, L=0.03) & 4.1\% & 1.6\% & 0.99 \\
         & \cellcolor{lightgray!30} Dataset3-LN (algo5, B=10k, L=0.03) & \cellcolor{lightgray!30} 4.1\% & \cellcolor{lightgray!30} 0.4\% & \cellcolor{lightgray!30} 0.99 \\
    \bottomrule
    \end{tabular}
    \caption{False positive and false negative rates for \lognormal{} detection at a threshold of (0.5) alongside with the AUC score. Algorithms are reported in Table \ref{tab:algorithms}, B stands for the budget used for the attack and L stands for the accepted $l^\infty$ distance. LN stands for \lognormal.}
    \label{tab:lognormal_results}
\end{table}

\section{Conclusions}\label{sec:conclu}
We tested \lognormal{} mutations on various benchmarks and extended it to continuous benchmarks, including fake detection tasks.%
Lognormal mutations perform well in some cases in the Nevergrad benchmark, especially on:  
(i) PBO, a classical discrete benchmark, and some other discrete benchmarks.
(2) In some continuous context, in particular in the most difficult scenarios such as many-objective, highly multimodal, low budget benchmarks, including e.g.~real world benchmarks in photonics. We note that other discrete algorithms adapted to the continuous case do perform well.

We note that discrete optimization methods are relevant in continuous optimization, when the prior (i.e.~the range of reasonable values for each variable) is important and excellent precision is impossible (Sections \ref{sec:nev} and \ref{selected}). Of course this can not compete with classical local optimization in terms of convergence rates for large budgets, but we show that it can be great for a low ratio budget/dimension.%
 Fake detection presents a complex challenge, with the difficulty of identification varying based on factors such as the type of data and the model used \citep{epstein2023online, sha2023fake}. We consider the setting in~\citep{wang2023dire}. We observe that \lognormal{} mutations in continuous settings, as well as other generic black-box optimization algorithms, are credible attack mechanisms, so that detecting attacks is necessary for a good defense.
 We observe a very poor transfer between the detectors of different types of attacks and in particular from the detection of classical attacks (such as SquareAttack) to our modified attacks (such as \lognormal), so that a good defense mechanism (aimed at improving the detection of manipulated images) must include the learning of diverse attack mechanisms.

Our most immediate further work is the combination of our fake detectors (no-box, SA, lognormal, and existing detectors such as \citep{ojha2023fakedetect} and \citep{fernandez2023stable}, into a single detector. We also add new attacks and defenses based on generic black-box optimization, and a third work is the investigation of \lognormal{} mutations (and other discrete algorithms) for other continuous problems.

\FloatBarrier

\bibliographystyle{tmlr}
\bibliography{bib}
\FloatBarrier

\appendix

\section{
Broader Impact Statement}

Detecting fake images is essential for preserving the quality of internet.
Our detectors do not provide certainties, only indications.

\section{Benchmarks: statistics}
\FloatBarrier
\begin{table}[ht]\centering\small
\begin{tabular}{|c|c|c|}
\hline
\rowcolor{lightgray!30} & Min & Max \\
\hline
Dimension & 1 & 20000 \\
\rowcolor{lightgray!30} Budget & 10 & 3000000 \\
Num-objectives & 1 & 6 \\
\rowcolor{lightgray!30} Noise dissymetries & False & True \\
Noise & False & True (many levels) \\
\rowcolor{lightgray!30} Number of blocks of variables & 1 & 16 \\ 
(with independent rotations) & & \\
Number of workers & 1 & 500 \\
\hline
\end{tabular}
\begin{tabular}{|c|c|}
\hline
\rowcolor{lightgray!30} Category & Benchmarks \\
\hline
Real-world, ML tuning     & Keras, Scikit-learn (SVM,\\
                          &  Decision Trees, Neural nets)\\
\rowcolor{lightgray!30} Real-world, not ML tuning & Crops, rockets, energy, fishing,\\
\rowcolor{lightgray!30}                           & photonics, game \\
Discrete & PBO, Bonnans, others \\
         & (includes: unordered variables) \\
\hline
\end{tabular}
\caption{\label{div}Diversity of the benchmarking platform used in our experiments.}%
\end{table}
Table \ref{div} presents the benchmarks in Nevergrad.
\FloatBarrier
\newpage
\section{Cases in which \lognormal{} is outperformed by NGOpt}
 \FloatBarrier
\begin{table}[ht]
\begin{tabularx}{\linewidth}{p{.35\textwidth}XXX}
\toprule
\multicolumn{4}{c}{Cases in which \lognormal{} does not outperform NGOpt}\\
\midrule
Problem & Rank of  LogNormal & Num algorithm & Rank of NGOpt \\
\midrule

yaonepenboundedbbob  & 7 & 86&  0 \\
yapenboundedbbob  & 8 & 87&  0 \\

yahdbbob  & 13 & 72&  7 \\
naivemltuning  & 14 & 21&  10 \\
reduced-yahdlbbbob  & 14 & 88&  1 \\
yamegapenboundedbbob  & 14 & 88&  8 \\
naive-seq-mltuning  & 15 & 23&  14 \\
seq-mltuning  & 17 & 23&  8 \\
yabbob  & 17 & 81&  11 \\
yaboxbbob  & 17 & 53&  7 \\
nano-naive-veryseq-mltuning  & 18 & 27&  17 \\
instrum-discrete*  & 23 & 123& 4 \\
mltuning  & 21 & 23&  9 \\
yaonepenparabbob  & 22 & 56&  11 \\
double-o-seven  & 24 & 39&  2 \\
yaboundedbbob  & 25 & 89&  4 \\
yapenparabbob  & 31 & 56&  13 \\
yatinybbob  & 41 & 89&  37 \\
zp-ms-bbob  & 41 & 58&  28 \\
ms-bbob  & 42 & 65&  29 \\
powersystems  & 42 & 50&  26 \\
yaparabbob  & 46 & 61&  6 \\
ultrasmall-photonics**  & 51 & 85&  28 \\
mldakmeans  & 57 & 70&  13 \\
\bottomrule
\end{tabularx}
\caption{\label{dag2}Benchmarks on which \lognormal{} performs weaker than NGOpt. NGOpt is a wizard, automatically choosing an algorithm in a big portfolio of algorithms: it performs vastly better than \lognormal{} for problems with noise (007 and PowerSystems, for which noise management is essential: \lognormal{} can do better for this when combined with Optimism in Front of Uncertainty for dealing with the noise as detailed in Section \ref{algos} and Fig. \ref{analy}), and for problems derived from BBOB in which high precision by continuous methods is possible.}
\end{table}
 
\FloatBarrier
\section{Black-box Optimization Algorithms}

\subsection{List of methods}\label{algos}
We briefly present the main BBO algorithms used in the present paper.

\begin{itemize}
    \item \Lognormal{} (the full name in Nevergrad is LogNormalDiscreteOnePlusOne, sometimes abbreviated as LDOPO), a $(1+\lambda)$~EA optimizer embedded with the self-adaptive \lognormal{} mutation. This optimizer is adapted from the corresponding discrete version \citep{Kruisselbrink2011lognormal} (see Section~\ref{sec:alg}), and details of the tested continuous optimizer are introduced in Section~\ref{sec:alg:extend}. 

    It is sometimes abbreviated as LDOPO.
    
    \item Lengler (DiscreteLenglerOnePlusone, sometimes abbreviated as  Lglr), namely the scheduled mutation rate in~\citep{relengler}. 

    \item Adaptive mutation rates (AdaptiveDiscreteOnePlusOne) come from \citep{lengler}. This is a $(1+1)$~EA with self-adjusting mutation rates. It increases the mutation rate $p$ by $F^s \cdot p$ when the obtained offspring is at least as good as the parent. Otherwise, $p$ is replaced by $p/F$. $F$ and $s$ are two constant parameters. The algorithm is extended to the continuous optimization following the strategy in Section~\ref{sec:alg:extend}. %
    \item Anisotropic discrete algorithms, which 
 uses self-adaptation with a mutation rate per variable. 
    \item NGOpt is a wizard developed by Nevergrad \citep{nevergrad}. It automatically chooses an algorithm from a portfolio of algorithms.
    \item CMA-ES, the Covariance Matrix Adaptation evolution strategy, is a well-known continuous optimization algorithm proposed by Hansen \citep{cma}. We consider CMA and DiagonalCMA \citep{diagcma} implemented by Nevergrad \citep{nevergrad} for comparison, and the latter is a modified version for high dimensional objective functions.
    \item Random Search (abbreviated RS) has been commonly applied as a baseline for algorithm comparison. In this paper, we sample new values for each variable uniformly at random. 
    
    \item MultiSQP, a combination of multiple sequential quadratic programming runs \citep{artelysSQP,nevergrad}.
    \item CMandAS2, a combination of several optimization methods (depending on the dimension and budget, e.g. several CMAs equipped with quadratic MetaModels combined in a bet-and-run\citep{betandrun} or the simple $(1+1)$ evolution strategy with one-fifth rule\citep{rechenberg1973}) implemented in \citep{nevergrad}.

    \item NGOptRW, the wizard proposed for real-world problems in \citep{nevergrad}; it used more DE and more PSO than NGOpt (as opposed to using CMA), and a bet-and-run\citep{betandrun}.
    \item Carola2, a chaining (inspired by \citep{ioh}) between Cobyla \citep{cobyla}, CMA \citep{cma} accompanied by a meta-model and sequential quadratic programming \citep{SQP}.

    \item HyperOpt \citep{hyperopt}, based on Parzen estimates.
    \item NgIoh variants, recent wizards co-developed by the Nevergrad team and the IOH team \citep{ioh}.

    \item CMA variants: besides DiagonalCMA \citep{diagcma}, we consider LargeDiagonalCMA (as DiagonalCMA, but with larger initial variance for the population).
    \item VastDE, basically DE \citep{DE} sampling closer to the boundary in bounded cases or with greater variance in unbounded contexts.
    \item SQOPSO, Special Quasi-Opposite PSO, which adapts quasi-opposite sampling~\citep{quasiopposite} to PSO.
    \item BAR4, a bet-and-run between \begin{itemize}\item quasi-opposite DE folowed by  BFGS with finite differences \citep{quasiopposite} and \item a CMA equipped with a meta-model followed by a sequential quadratic programming part. \end{itemize}
\end{itemize}

Other methods were run thanks to their availability in the Nevergrad framework; we refer to \citep{nevergrad} for all details. 

We use the terminology in the Nevergrad code, i.e., $(1+\lambda)$ optimization methods are derived from the $(1+1)$ code, and contain the suffix OnePlusOne even if $\lambda>1$.

\subsection{Modifiers of algorithms}\label{smooth}

\subsubsection{Modifiers dedicated to tensors}
When the Smooth operator is applied to a black-box optimization method in Nevergrad, periodically, it tries to replace the current best candidate $x$ by $Smooth(x)$.
If $Smooth(x)$ has a better loss value than $x$, then $x$ is replaced by $Smooth(x)$. 
$Smooth(x)$ is defined as a tensor with the same shape as $x$, with $Smooth(x)_i$ defined as $x_i$ with probability 75\%, and, otherwise, the average of the $x_j$ for $j$ the points at distance at most $1$ of $i$ in the indices of the tensor $x$.
Smooth means that this tentative smoothing is tested once per $55$ iterations, SuperSmooth once per 9 iterations; 
see Table \ref{tab:smooth}.
\begin{table}[t]\centering
\begin{tabular}{|c|c|}
\hline
\rowcolor{lightgray!30}Parameter & Value \\
\hline
Frequency of update & 1/2 (ZetaSmooth) \\
  (per iteration)   & 1/3 (UltraSmooth) \\
                    & 1/9 (SuperSmooth) \\
                    & 1/55 (Default smooth) \\
                    \hline
\rowcolor{lightgray!30}Size of smoothing window $s$ & $s=3$\\
\hline
\end{tabular}
\caption{\label{tab:smooth}Parametrization of the Smooth operator, which operates on a $s\times s$-window.}
\end{table}
 
\subsubsection{Modified dedicated to adversarial attacks}

The modifiers G and GSM used for arrays work as follows: 
\begin{eqnarray}
loss_{G}(x)&=& loss(0.03\times sign(x)) \nonumber\\
loss_{GSM}(x)&=& loss(0.03\times sign(convolve(x,k))) \nonumber
\end{eqnarray}
in the context of an image with values in $[0,1]$ and an amplitude $0.03$ in $l^\infty$.
$k$ refers to a Gaussian kernel of width $r/8$ with $r$ the width of the image.
 \FloatBarrier

\subsection{Other modifiers of algorithms proposed in Nevergrad}\label{optmodifiers}
 \FloatBarrier

There are other modifiers of algorithms proposed in Nevergrad and visible in prefixes, as follows:
\begin{itemize}

    \item The prefix SA (self-avoiding, referring to tabu lists) and the suffix Exp (referring to parameters related to simulated annealing) refer to add-ons for the discrete optimization methods.
    \item Recombining (R for short), which adds a two-point crossover \citep{holland} in an algorithm.
    \item Acceleration by meta-models: by default, MetaModel means CMA plus a quadratic meta-model, but an optimization method (DE, PSO or other) can be specified, and the meta-model can be a random-forest (RF) or a support vector machine (SVM) or a neural network (NN). The algorithm periodically learns a meta-model, and if the learning looks successful it uses the minimum of the MetaModel as a new candidate point.  %
\item Combination with Optimism in front of uncertainty: Nevergrad features bandit tools, which can be combined with other algorithms for making them compatible with noisy optimization.  For example, the optimistic counterpart of an algorithm $A$ performs an upper confidence bound method\citep{lairobbins} for choosing, between points in the search space that have already been used, which one should be resampled, and chooses a new point when the number of function evaluations exceeds a given function of the number of distinct points as specified in \citep{wam,nevergrad}.
\end{itemize}

 \FloatBarrier

\section{Significance in Nevergrad plots}\label{ranking}
 \FloatBarrier

In figures created by Nevergrad all dots are independently created. This means, for example,that dots obtained for budget 1000 are not extracted from truncations of runs obtained for budget 2000. This implies that the robustness of the rankings between methods can be deduced from the stability of curves: if the curve corresponding to algorithm A is always below the curve obtained for Algorithm B (in minimization) this means that Algorithm A performs robustly better than Algorithm B. For computing p-values, the probability that Algorithm A performs better than Algorithm B for the $k$ greatest budget values is at most $1/2^k$ under the null hypothesis that they have the same distribution of average loss values.
When computing ranks of algorithms as in Tables \ref{dag} and \ref{dag2}, Nevergrad proceeds as follows:
\begin{itemize}
\item Compute the average obtained loss $loss_{a,p,b}$ for each algorithm $a$ and each problem $p$ and each budget $b$ (averaged over instances).
\item Then the score of $a$ w.r.t algorithm $a'$ is $score_{a,a'}$, frequency at which $loss_{a,p,b}<loss_{a',p,b}$.
\item Then, given $n$ the number of algorithms, the score of $a$ is $score_a=\frac1n\sum_{a'} score_{a,a'}$ and Nevergrad provides the rank of $a$ for this score.
\end{itemize}

\FloatBarrier

\section{Comparisons of \lognormal{} variants}
 \FloatBarrier

\begin{figure}[ht]\centering
\includegraphics[width=.48\textwidth]{{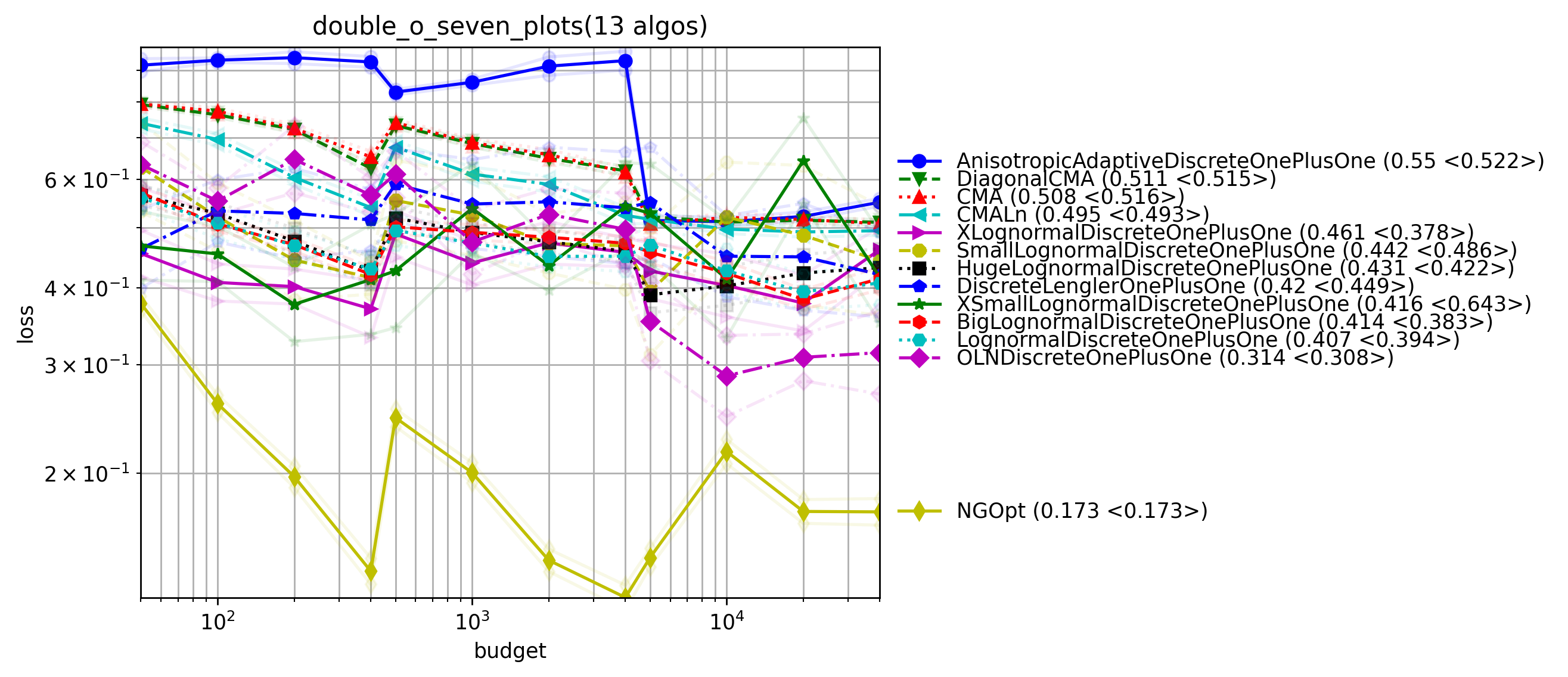}}
\includegraphics[width=.48\textwidth]{{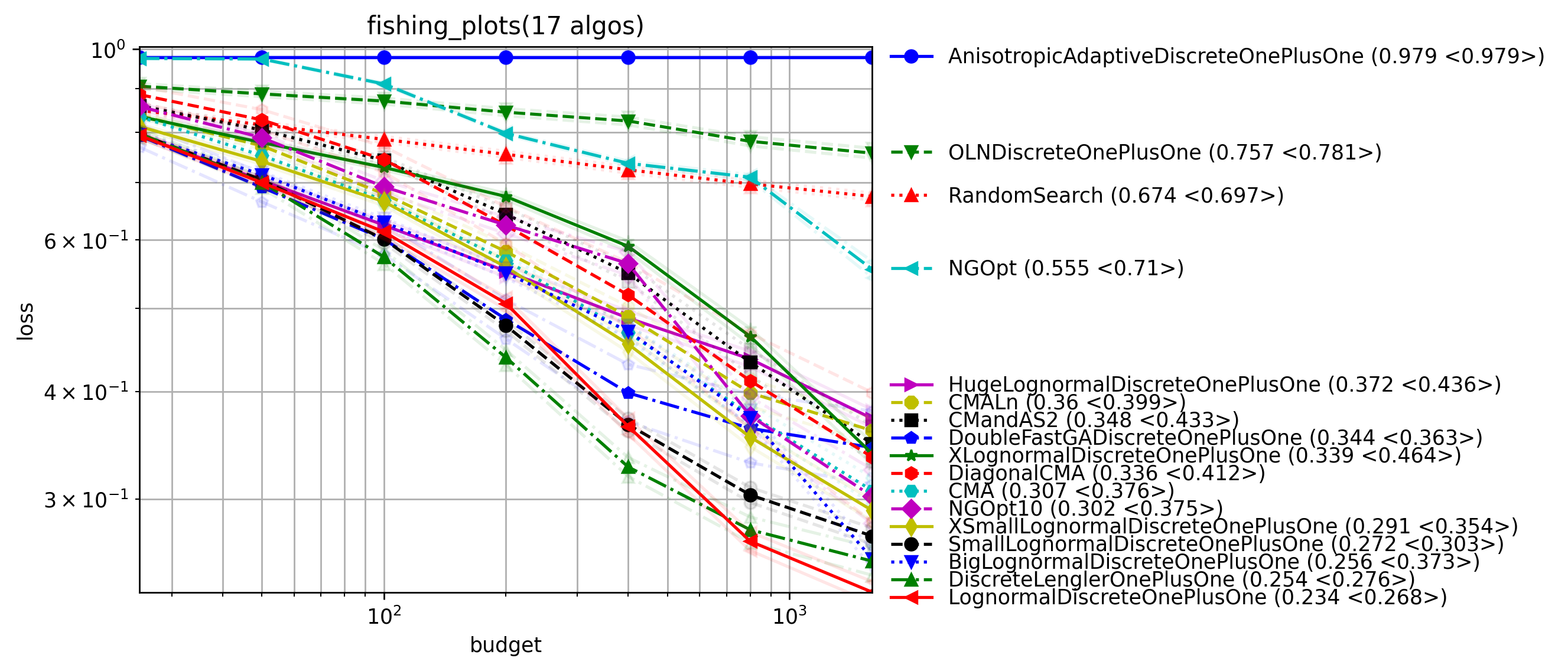}}
\includegraphics[width=.48\textwidth]{{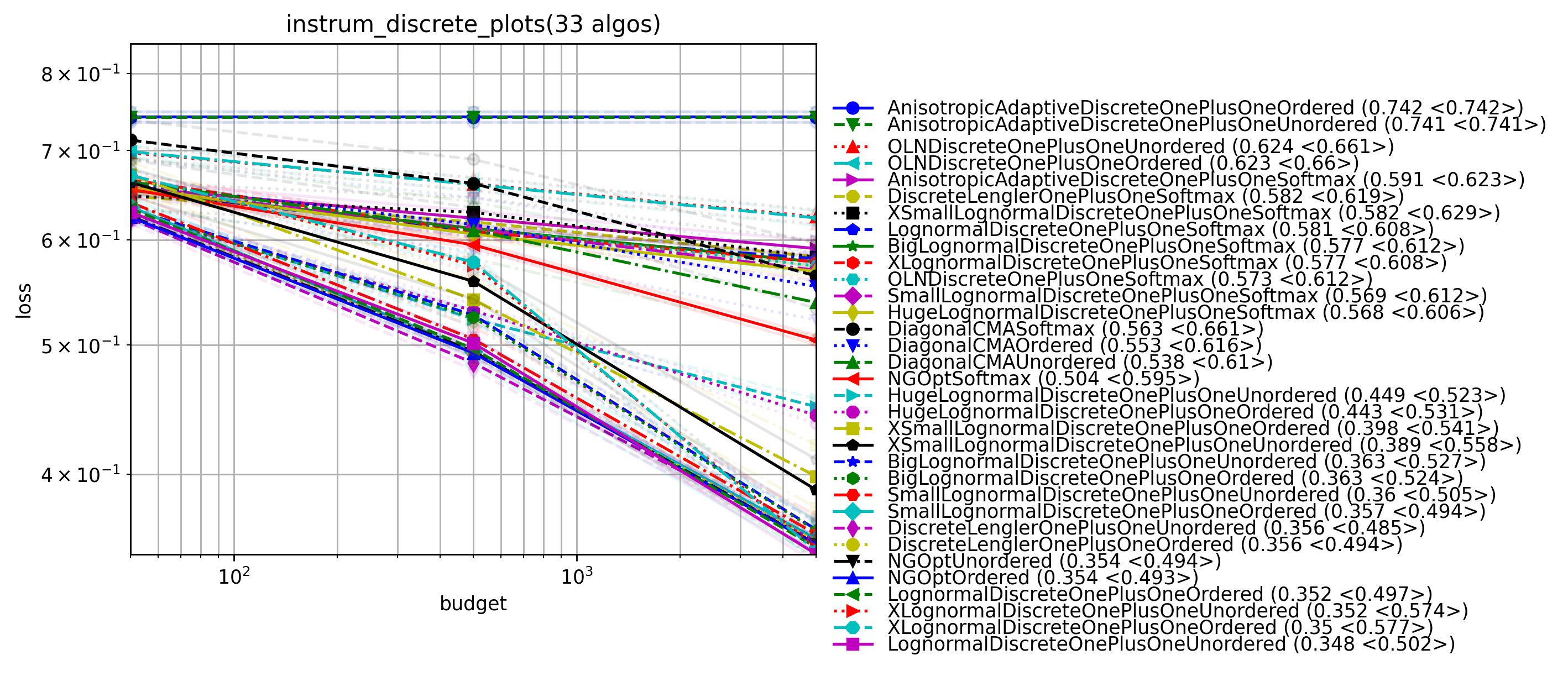}}
\includegraphics[width=.48\textwidth]{{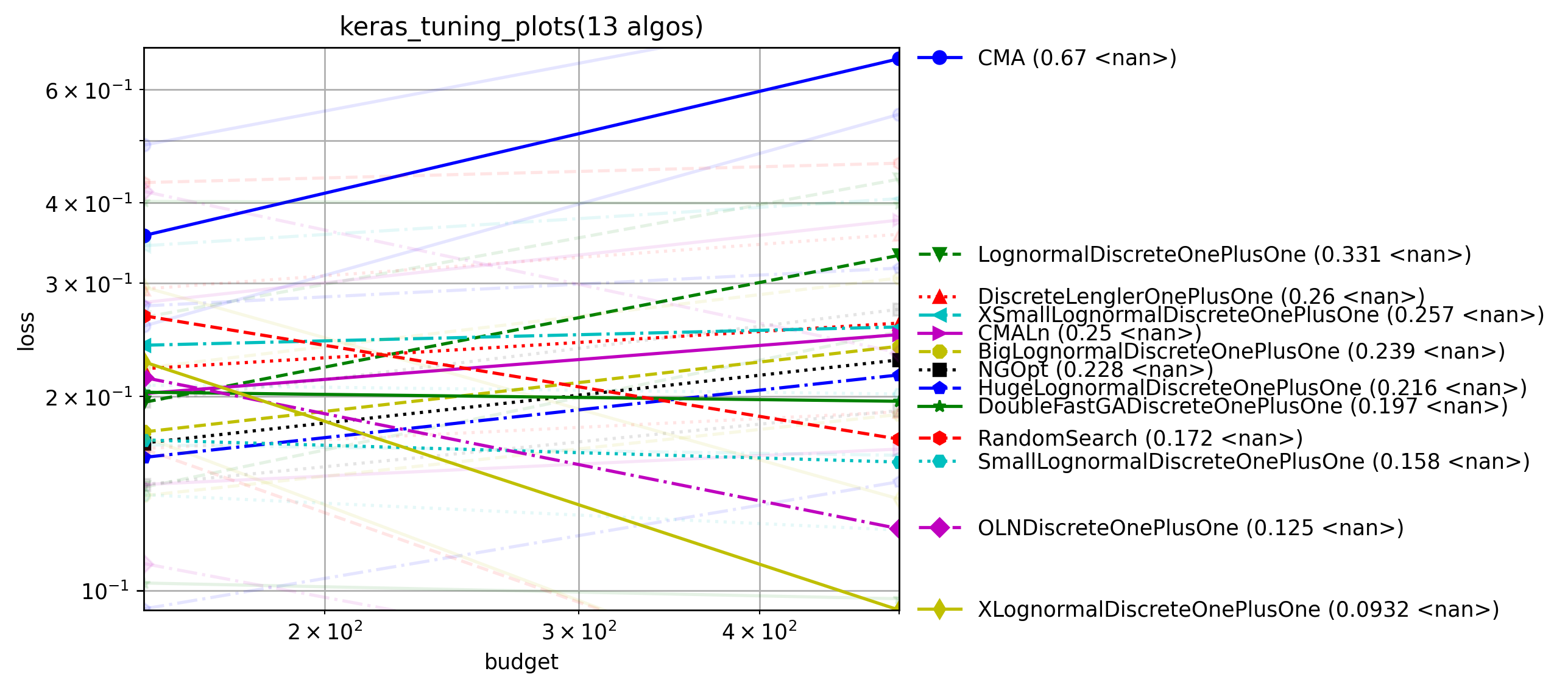}}
\includegraphics[width=.48\textwidth]{{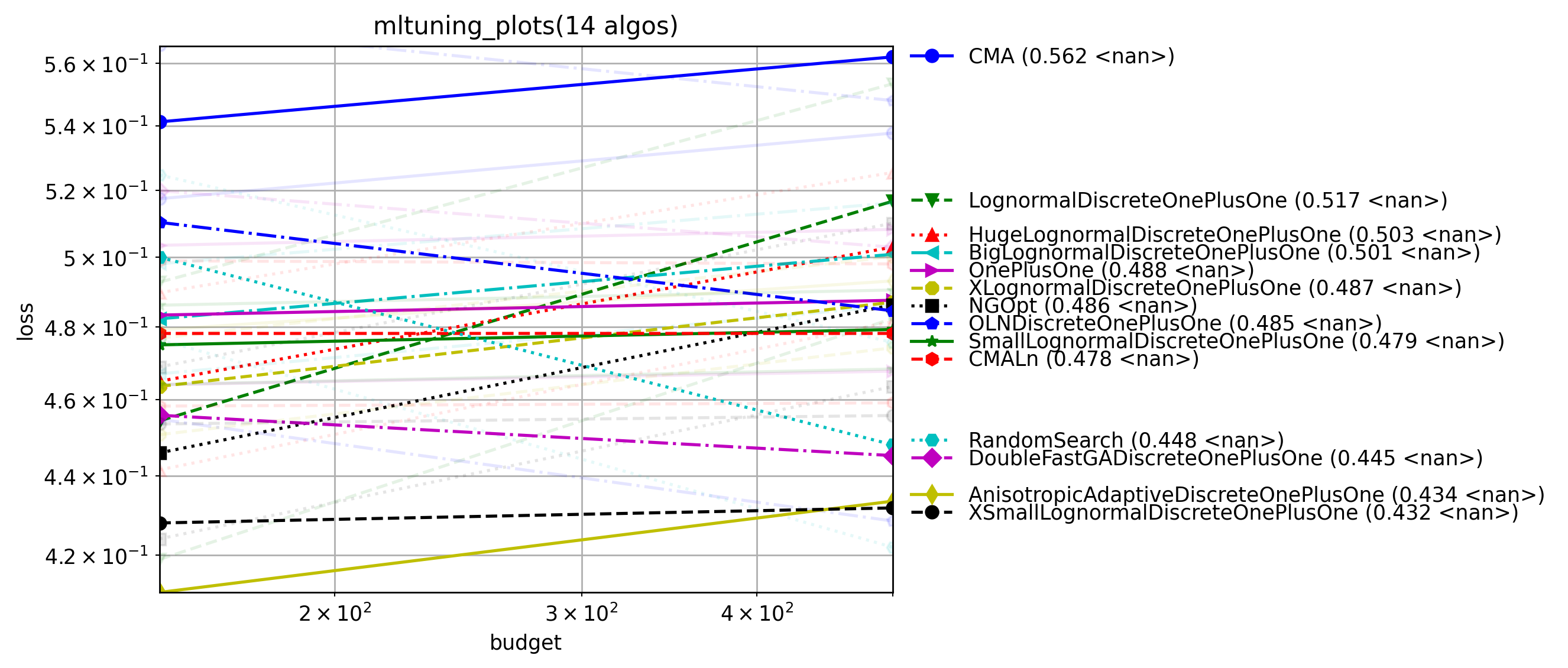}}
\includegraphics[width=.48\textwidth]{{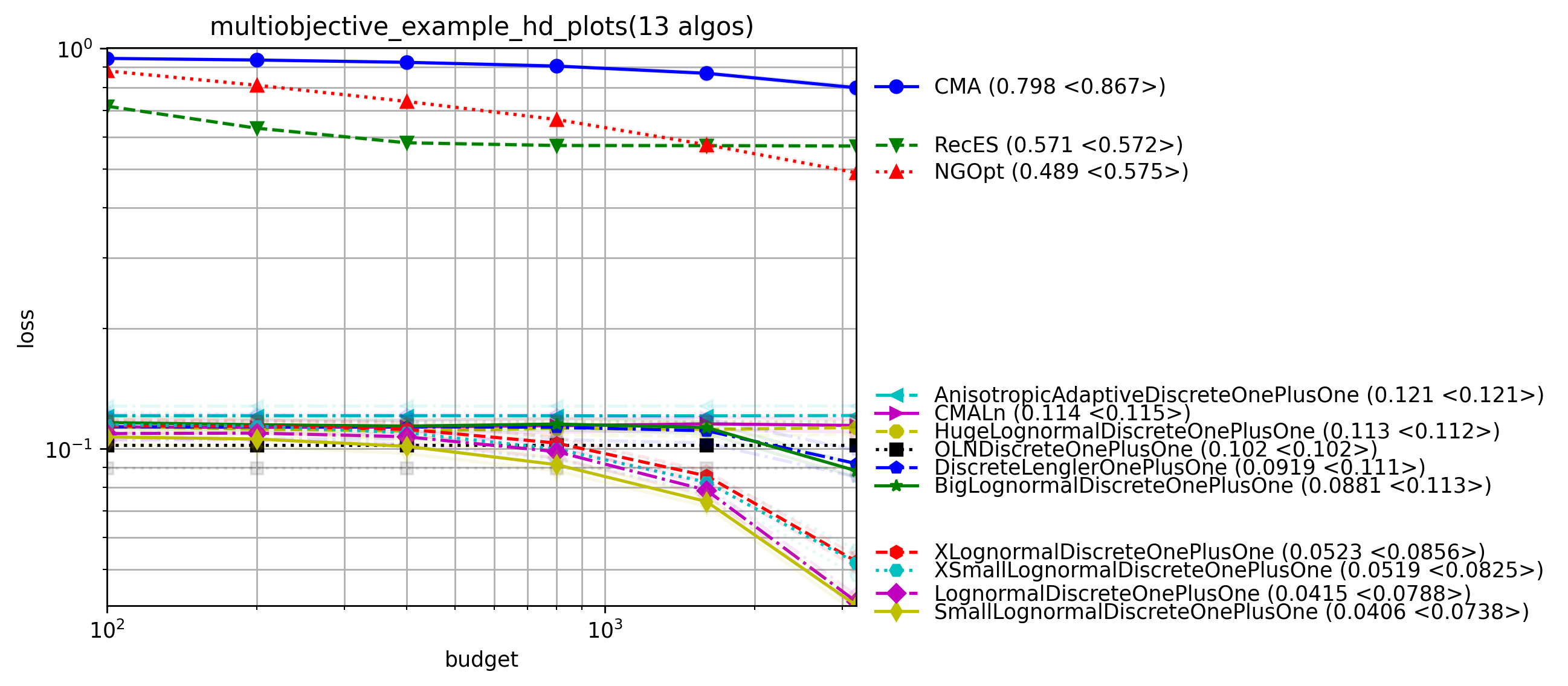}}
\includegraphics[width=.48\textwidth]{{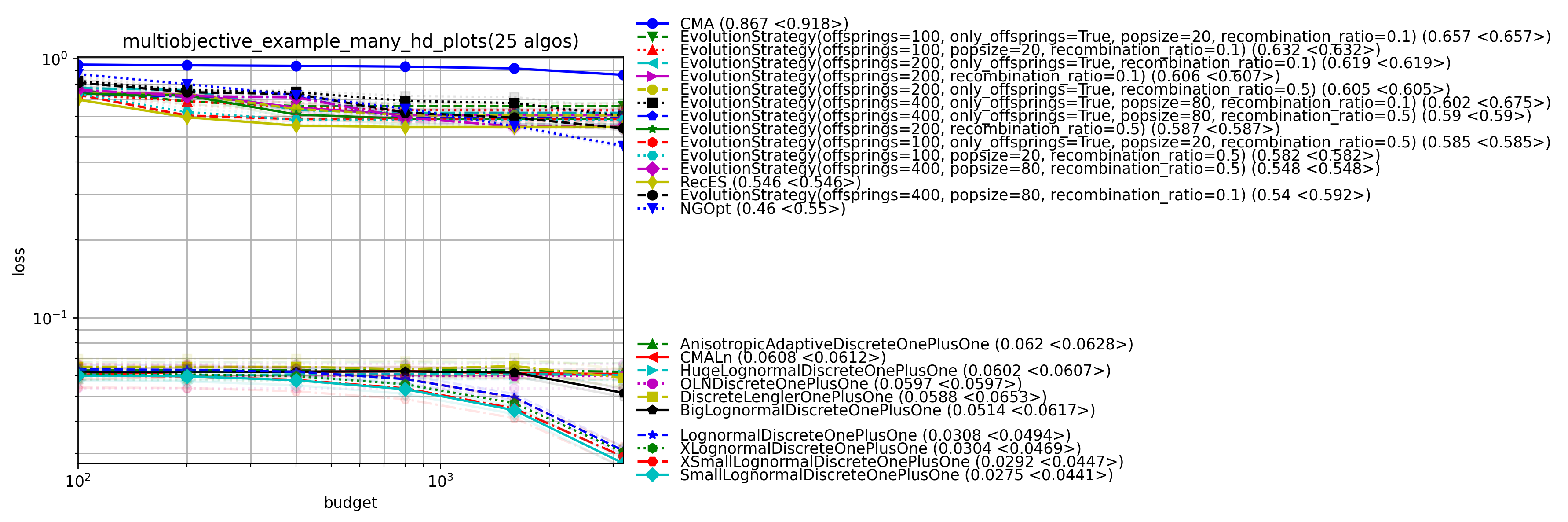}}
\caption{\label{analy}Analysis of variants of Lognormal mutations (1/2). The number between parentheses is the average score for the maximum budget and the number between brackets is the average score for the penultimate budget: as these two figures are obtained in completely independent runs the consistency between both shows the robustness/significance of the ranking (more details in Appendix \ref{ranking}). Unsurprisingly, the Optimistic variants (Section \ref{algos}) of \lognormal{} algorithms perform well for noisy optimization problems such as 007.} 
\end{figure}

\begin{figure}[t]\centering
\includegraphics[width=.48\textwidth]{{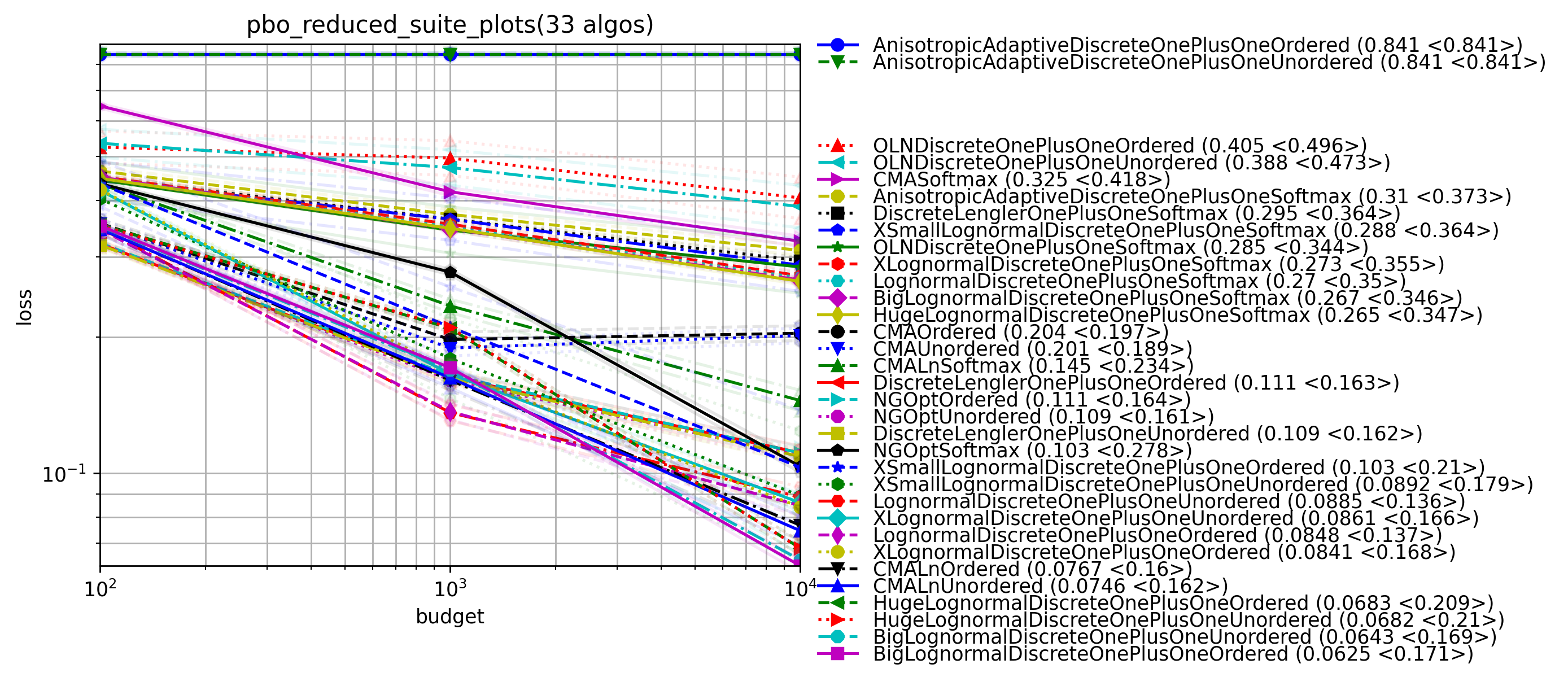}}
\includegraphics[width=.48\textwidth]{{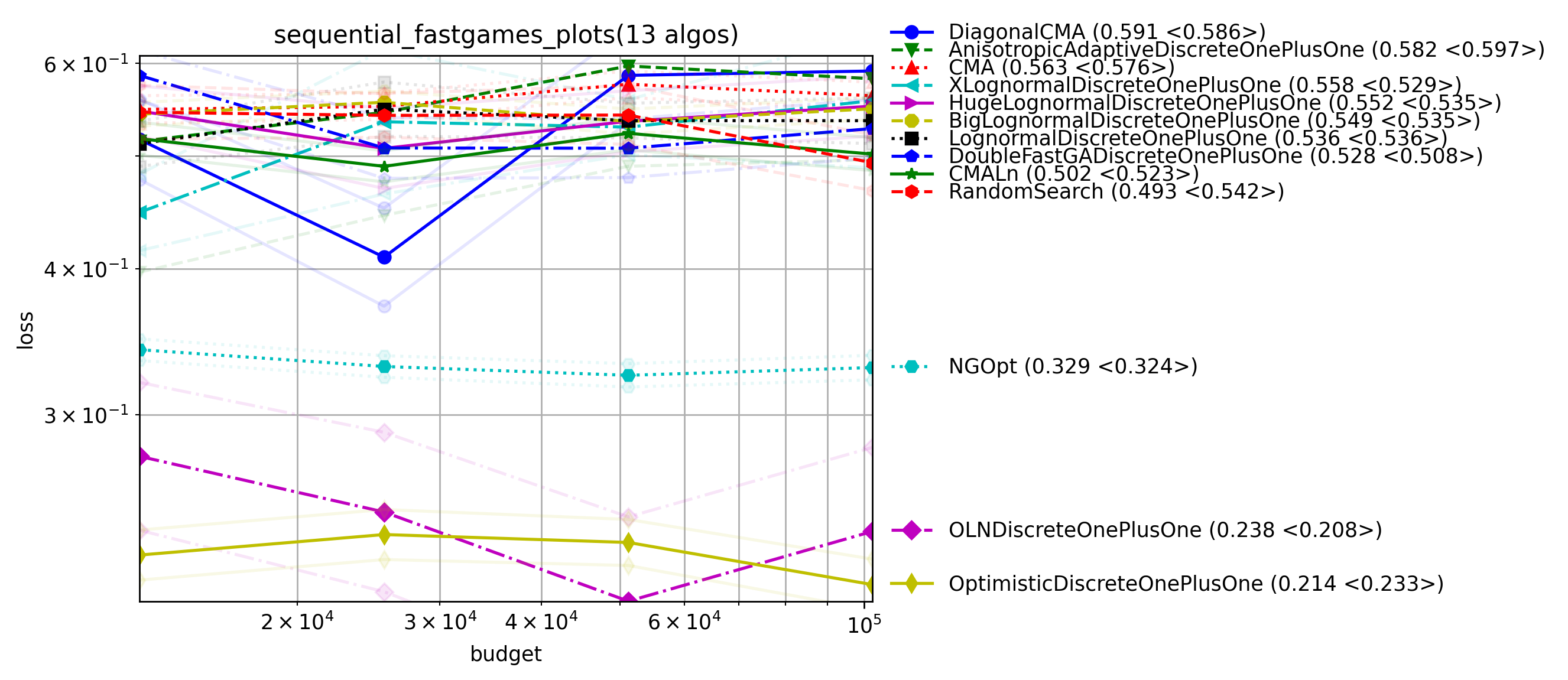}}
\includegraphics[width=.48\textwidth]{{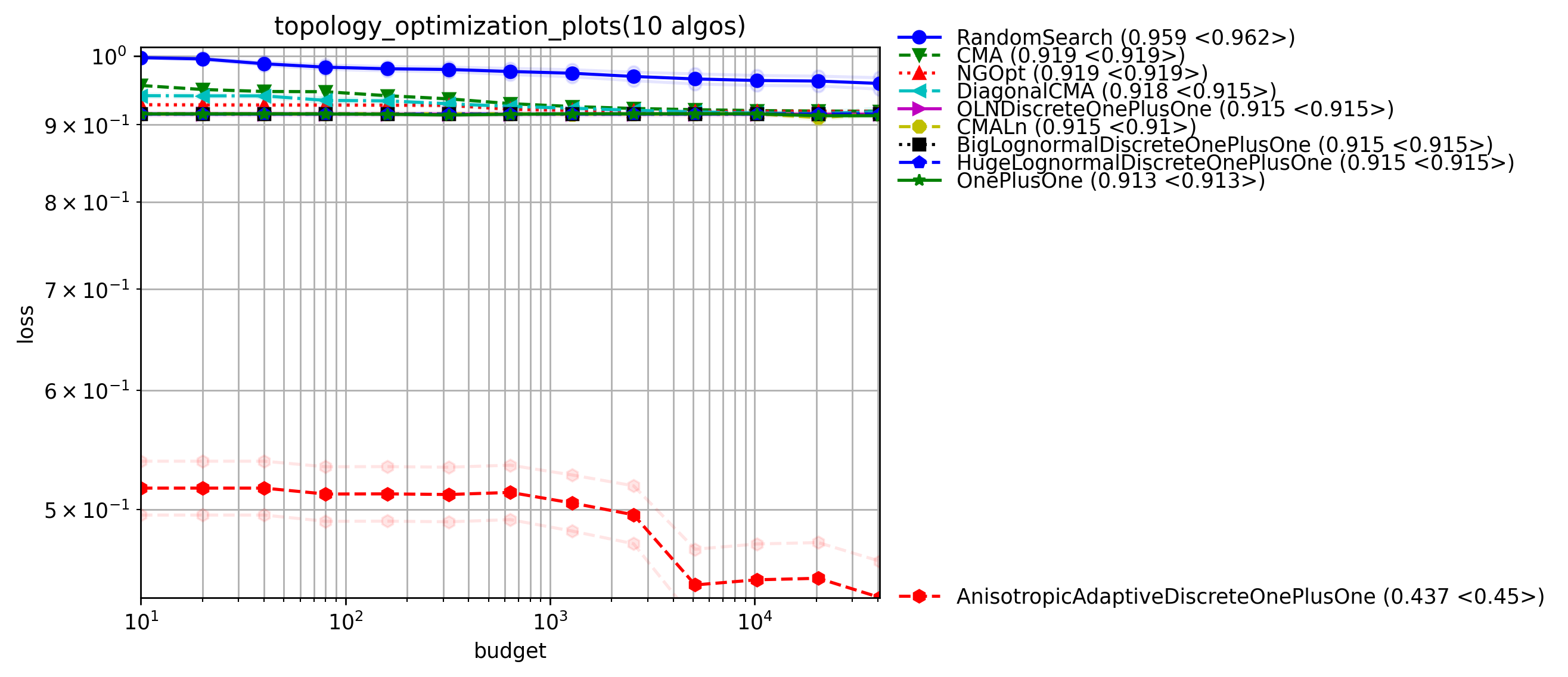}}
\includegraphics[width=.48\textwidth]{{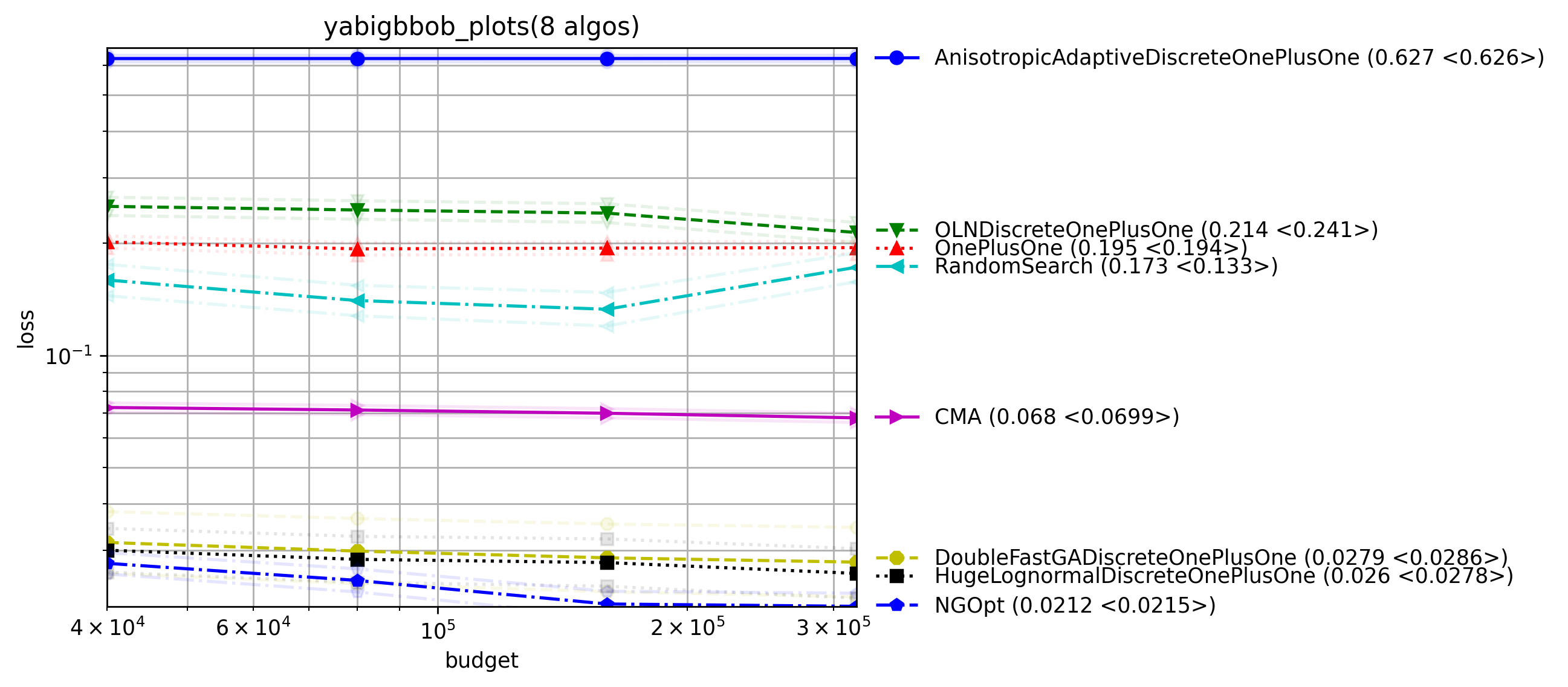}}
\includegraphics[width=.48\textwidth]{{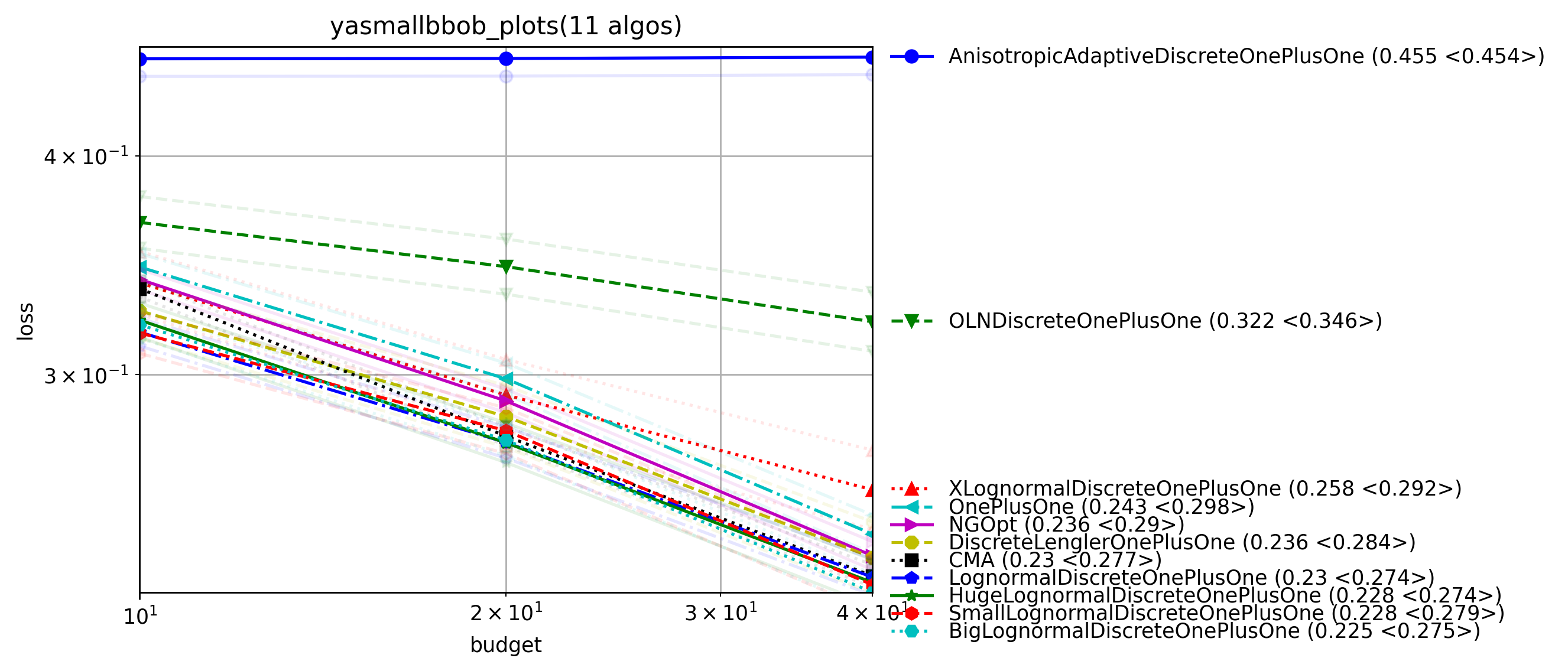}}
\includegraphics[width=.48\textwidth]{{dag_ln/topology_optimization_plots/xpresults_all.png}}
\includegraphics[width=.48\textwidth]{{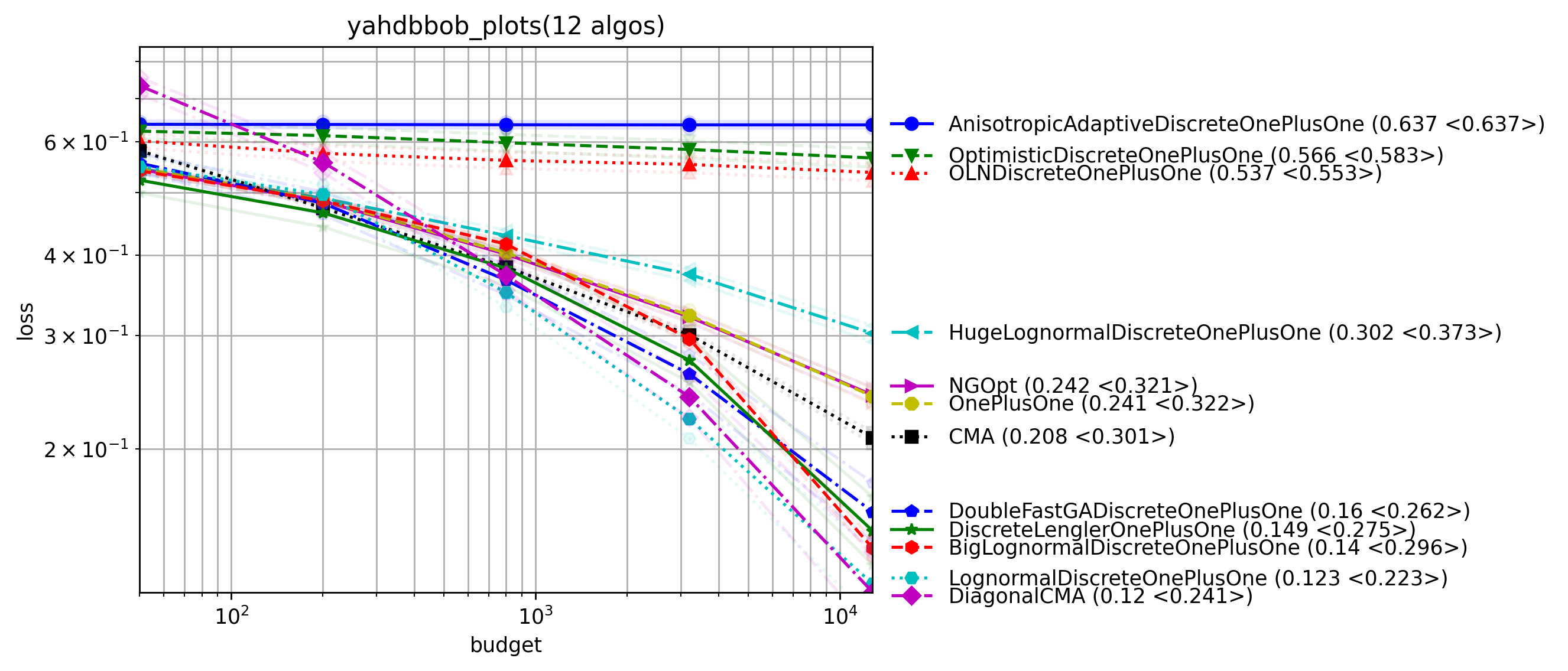}}
\caption{\label{analy2}Analysis of variants of Lognormal mutations (2/2). We observe that the default parametrization of LogNormal is essentially ok. Lognormal mutations are a serious competitor in some continuous problems when the prior search distribution used implicitly at initialization and mutation is good and the dimension is high (e.g., YAHDBBOB). For topology optimization, the variable-wise adaptive mutation rate  AnisotropicAdaptiveDiscreteOnePlusOne (see Section \ref{algos}) is excellent.}
\end{figure}

\label{appendix:compare-log-variants}
\Cref{analy,analy2} compare Lognormal variants on many problems. The optimistic variant is unsurprisingly good for noisy problems and topological problems seem to benefit from anisotropic mutations. Besides that, the standard Lognormal version is never very far from the optimum.
\FloatBarrier
\section{Parameters of the learning model}
\FloatBarrier
\begin{table}[ht]
\centering

\begin{tabular}{cc}
\toprule
\rowcolor{lightgray!30}
Hyperparameter & Value \\
\midrule
Type of model & SRnet\\
\rowcolor{lightgray!30}
learning rate & $10^{-4}$ \\
Number of epochs & 20 \\
\rowcolor{lightgray!30}
Hardware & 1 GPU \\
Training time & $\simeq$ 1 hour\\
\rowcolor{lightgray!30}
 & JPEG 0.2 prob at 96 quality\\
\rowcolor{lightgray!30} & Resize((256, 256))\\
\rowcolor{lightgray!30} & RandomHorizontalFlip(0.5)\\
\rowcolor{lightgray!30}\multirow{-4}{*}{Data Augmentation} & RandomCrop((224, 224))\\
\bottomrule
\end{tabular}

\caption{\label{tab:learner} Hyper-parameters of our learning model. We work on Dataset2 with each purifier (Dataset2-DP and Dataset2-IR) with parameter 0.1, which is split in 80\% train, 10\% test, 10\% validation.}
\end{table}
 \end{document}